\documentclass[journal,twoside,web]{ieeecolor}
\usepackage[dvipsnames, table]{xcolor}

\usepackage{generic}
\usepackage{amsmath,amssymb,amsfonts}
\usepackage{algorithm}
\usepackage{algorithmic}
\usepackage{graphicx}
\usepackage{epsfig}
\usepackage{booktabs}
\usepackage{textcomp}
\usepackage{adjustbox}
\usepackage{subcaption}

\makeatletter
\let\NAT@parse\undefined
\makeatother

\usepackage{hyperref}
\def\BibTeX{{\rm B\kern-.05em{\sc i\kern-.025em b}\kern-.08em
    T\kern-.1667em\lower.7ex\hbox{E}\kern-.125emX}}
\begin{document}
\title{XtraLight-MedMamba for Classification of Neoplastic Tubular Adenomas}
\author{Aqsa Sultana, \IEEEmembership{Student Member, IEEE}, Rayan Afsar, \IEEEmembership{Student Member, IEEE}, Ahmed Rahu, \textit{MD}, \\ Surendra P. Singh,\textit{ MD}, Brian Shula,  Brandon Combs, Derrick Forchetti, \textit{MD}, \\Vijayan K. Asari, \textit{PhD} \IEEEmembership{ Senior Member, IEEE}
\thanks{The authors would like to thank the South Bend Medical Foundation for providing access to the Neoplastic Tubular Adenomas dataset.}
\thanks{Aqsa Sultana and Vijayan K. Asari are with the Vision Lab and Dept. of Electrical and Computer Engineering, University of Dayton, Dayton, Ohio 45469, USA (e-mail: sultanaa3@udayton.edu; vasari1@udayton.edu). }
\thanks{Rayan Afsar is with the Vision Lab, University of Dayton, and Dept. of Computer Science, University of Georgia, Athens, Georgia, 30605. }
\thanks{Ahmed Rahu, MD, is a pathology resident, and Surendra P. Singh, MD, is a board-certified surgical and GI pathologist. Both are affiliated with The University of Toledo Medical Center, Toledo, Ohio, 43614}
\thanks{Brian Shula is with Honeywell International Inc., South Bend, Indiana, 46628}
\thanks{Derrick Forchetti, MD, a board-certified pathologist, and Brandon Combs are with the South Bend Medical Foundation, South Bend, Indiana, 46635 (email: dforchetti@sbmf.org).}
\thanks{The Neoplastic Tubular Adenoma (NPTA) dataset is available for download at \url{https://visionlab.udayton.edu/npta}}}

\maketitle

\begin{abstract}
Accurate risk stratification of precancerous polyps during routine colonoscopy screening is a key strategy to reduce the incidence of colorectal cancer (CRC). However, assessment of low-grade dysplasia remains limited by subjective histopathologic interpretation. Advances in computational pathology and deep learning offer new opportunities to identify subtle, fine morphologic patterns associated with malignant progression that may be imperceptible to the human eye. In this work, we propose XtraLight-MedMamba, an ultra-lightweight state-space–based deep learning framework to classify neoplastic tubular adenomas from whole-slide images (WSIs). The architecture is a blend of a ConvNeXt-based shallow feature extractor with parallel vision mamba blocks to efficiently model local texture cues within global contextual structure.  
An integration of the Spatial and Channel Attention Bridge (SCAB) module enhances multiscale feature extraction, while the Fixed Non-Negative Orthogonal Classifier (FNOClassifier) enables substantial parameter reduction and improved generalization. The model was evaluated on a curated dataset acquired from patients with low-grade tubular adenomas, stratified into case and control cohorts based on subsequent CRC development. XtraLight-MedMamba achieved an accuracy of 97.18\% and an F1-score of 0.9767 using approximately 32,000 parameters, outperforming transformer-based and conventional Mamba architectures, which have significantly higher model complexity and computational burden, making it suitable for resource-constrained areas.
\end{abstract}

\begin{IEEEkeywords}
Digital pathology, tubular adenoma, Colorectal Cancer, parallel vision mamba, state space models, ConvNeXt, lightweight deep learning, whole-slide images
\end{IEEEkeywords}

\section{Introduction}
\label{sec:introduction}

\begin{figure*}[h]
    \centerline{%
        \includegraphics[width=0.9\linewidth,height=7cm]{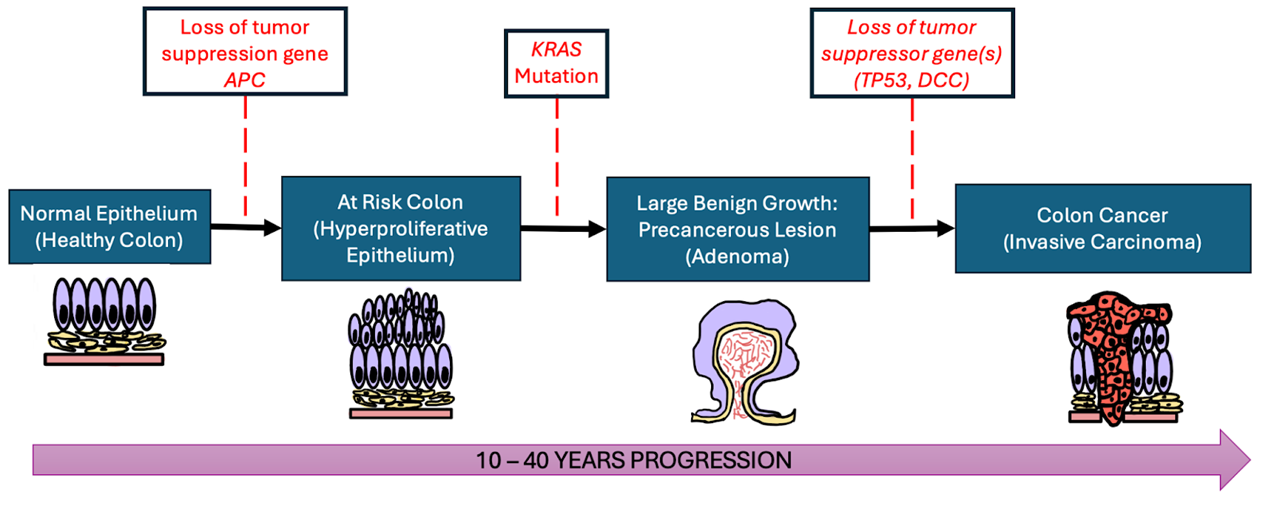}%
    }
    \caption{Schematic illustrating the concept of the adenoma–carcinoma sequence described by Fearon and Vogelstein \cite{genetic-alterations}. Together, these observations suggest that the central idea is that colorectal cancer develops in a step-wise progressive manner, starting from the precursor lesions (adenomatous polyps), eventually progressing to invasive carcinoma. Early loss of APC is associated with hyperproliferative epithelium, increasing the risk of colonic polyp formation. This is followed by an activating mutation in KRAS, which itself promotes the formation of adenomatous polyps. Subsequent loss of tumor suppressor genes, including TP53 and DCC, contributes to tumor progression and eventual neoplastic progression towards invasive colorectal adenocarcinoma \cite{genetic-alterations}. This step-wise progression from adenoma(s) to cancer typically takes over ten years to progress \cite{epidemiology}.
}
    \label{fig1}
\end{figure*}

\IEEEPARstart{C}{olorectal} cancer (CRC) remains a major public-health burden worldwide. It is the third most commonly diagnosed cancer and the second leading cause of cancer-related mortality in the United States \cite{ACS2025},\cite{cancer-statistics}.
Colonic polyps are raised protrusions of colonic mucosa, comprising several histologic subtypes with differing biological behavior. Among these, adenomatous polyps represent a major category of premalignant lesions arising from the neoplastic proliferation of colonic glands. Neoplasia refers to abnormal clonal cell growth that persists in the absence of normal regulatory signals. Although these types of polyps are typically benign, this autonomous proliferative capacity is why adenomatous polyps are considered precancerous lesions and can progress to cancer via the adenoma-carcinoma sequence in a stepwise manner \cite{Rosai_2011},\cite{Levin2008_CRC_Screening_Guideline},\cite{genetic-alterations}  as shown in Fig. \ref{fig1}. 
Over decades, the classical adenoma–carcinoma sequence has established adenomatous polyps as a central biological intermediate in colorectal tumorigenesis, forming the basis for modern CRC screening and surveillance strategies \cite{genetic-alterations},\cite{Force_2021}.

Amongst the adenomatous polyps are subtypes. Tubular adenomas (TAs) represent the most prevalent subtype encountered in routine clinical practice and are therefore of particular clinical importance. Adenomatous polyps are histologically classified as low-grade or high-grade based on the degree of dysplasia (abnormal tissue architecture). In general, high-grade dysplasia is a characteristic predictor of advancement to colorectal carcinoma \cite{Rosai_2011}. 
The majority of TAs detected via screening colonoscopy and histopathological examination typically demonstrate low-grade dysplasia, in which malignant potential is low-risk \cite{Rosai_2011},\cite{epidemiology}. Current histopathologic assessment relies primarily on subjective visual interpretation, which may be insufficient to identify subtle or spatially distributed morphologic features associated with long-term development of CRC in this ostensibly low-risk population.
Epidemiologic trends further highlight the importance of improving risk stratification at the precursor lesion stage \cite{ACS2025},\cite{Levin2008_CRC_Screening_Guideline}. Despite widespread screening initiatives among adults aged 45+ that have reduced the overall incidence of CRC, important age-related disparities persist; in recent years, incidence has continued to rise among younger individuals despite overall declines \cite{ACS2025},\cite{Force_2021}. Historical and recent data continue to highlight the clinical relevance of early detection and the limitations of current risk assessment strategies based on coarse histologic categorization. 
Despite the success of screening programs in reducing CRC incidence and mortality, a significant unmet clinical need remains in accurately stratifying risk among patients with low-grade tubular adenomas (TAs). Most individuals diagnosed with TAs are classified as low risk and managed conservatively, yet a subset will continue to later develop high-grade dysplasia, neoplasia, and/or CRC. Continued progress in CRC prevention will therefore require improved tools capable of extracting objective, quantitative information from precursor lesions beyond what is achievable through conventional histopathologic evaluation alone.
The increasing adoption of digital pathology has transformed histopathology into a data-rich imaging modality through the routine acquisition of high-resolution whole-slide images (WSIs). When coupled with advances in machine learning, digital pathology offers the potential to enhance prophylactic efforts by enabling large-scale, quantitative analysis of tissue morphology and by uncovering subtle patterns imperceptible to human observers \cite{LITJENS201760},\cite{MADABHUSHI2016170}. However, the high resolution, heterogeneity, and weak labeling of WSI data introduce substantial computational challenges, necessitating efficient, scalable analytical frameworks.

Recent advances in computational pathology and deep learning architectures have enabled automated feature extraction from WSIs, allowing models to learn hierarchical representations of tissue patterns directly from histological data without manual intervention. With early efforts driven by convolutional neural network (CNN)-based models \cite{oshea2015-introductionconvolutionalneuralnetworks}, \cite{he_deep_2016}, \cite{rcnn}, \cite{tan2021efficientnetv2smallermodelsfaster}, and subsequently transformer-inspired models \cite{DBLP:journals/corr/abs-2103-14030}, \cite{DBLP:journals/corr/abs-2010-11929}, strong performance has been achieved by modeling local textural features and global context through self-attention mechanisms, respectively.
While effective, CNNs have a limited receptive field, constraining their ability to model long-range spatial relationships that are critical for understanding broader tissue architecture. Whereas transformers incur computational overhead due to their quadratic complexity, which poses challenges for WSI analysis tasks.

Most recently, state space models (SSMs), have introduced new opportunities for efficient representation learning in large-scale visual data \cite{gu2022efficientlymodelinglongsequences,gu2024mambalineartimesequencemodeling}. Vision Mamba, \cite{zhu2024visionmambaefficientvisual,rahman2024mambavisioncomprehensivesurvey,Sultana_Abouzahra_Asari_Aspiras_Liu_Sudakow_Cooper_2025}, an SSM-based architecture, enables modeling of short- and long-range spatial dependencies with linear computational complexity, all while maintaining favorable computational and memory characteristics, making it well-suited for WSI analysis. In this context, the application of modern, computationally efficient architectures to digital pathology represents a critical step toward advancing cancer prevention through improved risk stratification of precursor lesions. 

Conventional risk stratification methods categorize adenomas by size, histologic subtype, and grade of dysplasia, potentially overlooking finer-scale tissue patterns that may reflect underlying biological behavior. TAs with low-grade dysplasia are often regarded as biologically homogeneous, despite evidence that not all precancerous lesions progress to malignant disease. Subtle morphologic cues and variations, such as nuclear architecture, glandular organization, and epithelial maturation, may be difficult for human observers to consistently quantify but can be captured by deep learning models. This paper proposes XtraLight-MedMamba, an ultra-lightweight architecture incorporating ConvNext blocks, parallel Mamba-based layers, Spatial and Channel Attention Bridge (SCAB) and  Fixed Non-Negative Orthogonal Classifier (FNOClassifier).  ConvNeXt blocks extract shallow features from the input image and Mamba layers enhances and identifies subtle morphologic pattern cues through the modeling of both short- and long-range spatial dependencies. Then, SCAB module refines the intermediate feature representations. Whereas, FNOClassifier is a parameter-efficient classification module that classifies neoplastic TA features into ``case'' and ``control'' cohorts. The task-specific integration of these components into a compact architecture enables strong representation of both local glandular morphology and broader contextual patterns while maintaining an exceptionally low parameter count. Experimental results demonstrate that XtraLight-MedMamba outperforms previously established models, as detailed in subsequent sections.

Our main contributions include:
\begin{itemize}
    \item A Mamba-based ultra-lightweight architecture for classification of neoplastic tubular adenomas: XtraLight-MedMamba, combining the strengths of ConvNeXt blocks, mamba modules in parallel layers, a spatial and channel attention bridge module, and a fixed non-negative orthogonal classifier within a unified framework.  
    \item A unique deep learning architectural strategy  integrating (1) ConvNeXt blocks as a shallow feature extractor, acting upon design principles of Vision Transformers \cite{dosovitskiy_image_2021} (ViT) such as depthwise separable convolutions, Layer Normalization, and MLP-style channel expansion to efficiently capture local morphologic patterns while preserving the spatial inductive bias of CNNs. (2) Parallel Vision Mamba (PVM) layers are included to model both short- and long-range spatial dependencies in parallel by branching out the features for state-space sequence modeling, each with channel \(C/4\), enabling efficient global contextualization of morphologic features with linear computational complexity. (3) Multiscale feature fusion is enhanced by integration of SCAB modules by emphasizing relevant informative spatial regions and channel-wise responses. (4) FNOClassifier is incorporated to reduce redundant parameters deeper in the layers of the model while improving generalization and training stability through structured, orthogonal feature-to-class projections. This task-specific integration of these components into a compact architecture is tailored for subtle histopathologic pattern recognition.
    \item Evaluation of the risk stratification performance of XtraLight-MedMamba on our newly introduced Neoplastic Tubular Adenoma (NPTA) dataset, a curated case-control dataset derived from WSIs of low-grade TAs.
\end{itemize}

The proposed framework achieves strong classification performance with very low model complexity, yielding a favorable performance efficiency trade-off. By focusing on morphologic features critical for tissue characterization, XtraLight-MedMamba aims to improve the accuracy of neoplastic tubular adenoma classification by capturing subtle cellular and complex architectural patterns from case and control cohorts.

\section{Related Work}
Computational analysis of histopathology images has
become a key research focus for improving CRC diagnosis and evaluation, driven by advances in computer vision. Early approaches relied on handcrafted algorithms that focused on specific image features to distinguish tissue types. Hamilton et al. \cite{hamilton_automated_1997} introduced image texture analysis using co-occurrence matrix features and low optical density feature counts for classifying normal versus dysplastic tissues. This method automated the localization of dysplastic fields in colorectal histology. Subsequently, Kalkan et al. \cite{kalkan_automated_2012} introduced an automated diagnosis of colorectal cancer from a whole biopsy slice that combined textural and structural features, using a two-level classification scheme. First, individual patches were classified as adenomatous, inflamed, cancerous, or normal using a k-nearest neighbors classifier. Afterwards, the slice-level information obtained from the patch distribution was then used to classify the slices using a logistic linear classifier. Another method proposed fully automated CRC grading from histology images \cite{sengar_grading_2016}. The glands were first segmented automatically using an intensity-based thresholding approach, and the images were then classified as benign healthy, benign adenomatous, moderately differentiated malignant, or poorly differentiated malignant with a support vector machine classifier. While these methods were moderately effective for their respective tasks, they did not achieve the accuracy required to be viable tools for pathologists to detect and grade colorectal cancer from WSIs.

\subsection{Deep Learning for Colorectal Cancer Histopathology}
With the advent of deep learning, classical approaches have been largely replaced for automated CRC classification from histopathology images. CNNs were among the first deep learning architectures widely applied to histopathological image analysis, and they have been reviewed extensively alongside classical machine learning and CNN-based methods for classification, detection, and segmentation of relevant tissue \cite{zhou_comprehensive_2020}. Gastrointestinal-related reviews further explored applications of deep learning for CRC detection and prognosis from histology slides \cite{kuntz_gastrointestinal_2021}. Consequently, deep learning has become a major application in CRC histopathology, as evidenced by a systematic review of deep-learning-based CRC diagnosis \cite{davri_deep_2022}. A transfer learning framework for classifying CRC histology images with sparse WSI annotations was proposed by Ben Hamida et al. \cite{ben_hamida_deep_2021} and evaluated using a benchmark of multiple pre-trained CNN models. In their experiments, patch-level classification on their AiCOLO dataset was performed with an 18-layer ResNet \cite{he_deep_2016} model. In addition, a multi-step segmentation model (UNet \cite{ronneberger_u-net_2015}, and SegNet \cite{badrinarayanan2016segnetdeepconvolutionalencoderdecoder}) was employed to generate semantic maps to identify abnormal regions on the slides. 
Meanwhile, Abdulrahman et al. \cite{abdulrahman2025optimized} proposed an ensemble model combining EfficientNetV2 and DenseNet for binary classification of malignant versus benign colorectal tissue from WSIs using a custom Bahrain hospital dataset. 

Many models are trained on fixed-size patches of WSIs, where the patch-level predictions can be aggregated to obtain slide-level classification \cite{wang_accurate_2021,steimetz_deep_2025}. This strategy is computationally tractable for gigapixel-sized WSIs, applicable to a wide range of architectures, and has achieved strong performance across tasks, including cancer detection and tumor grading. For example, Wang et al. \cite{wang_accurate_2021} used a weakly supervised, transfer-learned Inception-v3 for CRC detection and achieved slide-level classification accuracy on par with that of human pathologists. Paing and Pintavirooj \cite{paing_adenoma_2023} proposed a ResNet-based architecture that applies a Fast Fourier Transform to a ResNet50 backbone for tumor grading, achieving similarly high performance. This model used cross-feature fusion to fuse local-scale spatial convolutions with global-scale Fourier convolutions. Another study by Steimetz et al. \cite{steimetz_deep_2025} used an ImageNet-pretrained ResNet-34 model to classify between low-grade and high-grade dysplasia.
Likewise, Zhou et al. \cite{zhou_construction_2024} introduced a deep learning-based tumor risk signature (TRS) approach, using an ensemble of VGG19, ResNet50, and DenseNet21 models for the detection of stage III colorectal cancer. The model first segmented nine tissue types in WSIs, and subsequently, the tumor features were extracted to fit a Cox proportional hazards model. Similarly, DenseNetV2 \cite{dragomir_quantitative_2025} was fine-tuned using the HALO image analysis platform to automatically detect the six morphotypes. The trained model was then applied to 644 sections from 161 cases to quantify morphotype areas. Furthermore, A lightweight, non-pretrained CNN \cite{li_lightweight_2025} was proposed for the detection and visualization of multiclass colon tissue. It was integrated with a parametric Gaussian-distribution-based data-cleaning strategy to remove outliers and improve data quality. Another study employed a few-shot learning approach \cite{li_few-shot_2024} that combined transfer learning and contrastive learning to classify CRC histopathology into benign and malignant categories. The model comprised modules for feature extraction, dimensionality reduction, and classification, and was trained using contrastive and cross-entropy losses. Subsequently, an ensemble deep learning model that combined the watershed algorithm to enhance glandular segmentation in colon histopathology was proposed by Roy et al. \cite{roy_revolutionizing_2024}. This approach employed a UNet-based CNN, a weighted ensemble network that integrated DenseNet169 via augmentation, InceptionV3, and EfficientB3 as the backbone. 
Despite their success, conventional CNNs are inherently limited by the locality of convolutional operations.

To tackle this issue, Transformer-based architectures, enabled by the introduction of the Vision Transformer (ViT) \cite{dosovitskiy_image_2021}, have been explored to better capture long-range spatial relationships within colorectal tissue \cite{sathyanarayana_colovit_2025,tp_deepcpd_2024,qin_colorectal_2024}, and for the detection of higher-order structures within CRC \cite{chen_predicting_2024}. Transformer-based architectures relied heavily on self-attention to capture such subtle patterns, which exhibit quadratic scaling \cite{gu2022efficientlymodelinglongsequences,keles_computational_2023}, making them difficult to scale efficiently.

 ViT \cite{dosovitskiy_image_2021} converted the input image into a set of patch-level embeddings, flattened them, and processed them like a sequence of tokens similar to Natural Language Processing (NLP) \cite{vaswani2023attentionneed}. Using multi-head self-attention, the patches were then processed to capture information from across the image.
 While effective, ViT typically benefited from large datasets and high computational resources. To make transformers more scalable for images, Swin Transformer \cite{liu_swin_2021}, also known as 
 \textbf{S}hifted \textbf{Win}dow Transformer was introduced, as an efficient variant of Transformers. Swin Transformers employed a hierarchical design, where attention was computed using non-overlapping local windows to minimize computational overhead. To maintain efficiency and prevent leakage of information, the window partitions were shifted between layers, allowing information to flow across neighboring windows.

In colonoscopy imaging, ColoViT \cite{sathyanarayana_colovit_2025} was introduced as a hybrid of EfficientNet and ViT for advanced colon cancer detection. Here, EfficientNet extracted local features, whereas ViT captured global contextual information in colonoscopic images. Meanwhile, an efficient deep learning method, DeepCPD \cite{tp_deepcpd_2024}, was introduced to classify colonoscopic images into polyp versus non-polyp and hyperplastic versus adenoma. The model was a combination of a transformer and a Linear Multihead self-attention (LMSA) mechanism with data augmentation. In contrast, for histology, a hybrid of Swin Transformer for feature extraction and a skip-feedback connection with UNet was proposed to improve the model's multi-level feature extraction capabilities, enabling end-to-end recognition of colorectal adenocarcinoma tissue images \cite{qin_colorectal_2024}. Subsequently, Chen et al. \cite{chen_predicting_2024} proposed DiNAT-MSI, an algorithm that incorporated the Dilated Neighborhood Attention Transformer (DiNAT) to enhance global context recognition and expand receptive fields, while avoiding fully quadratic global attention. Likewise, Transformer-Based Self-Supervised Learning and Distillation \cite{li_transformer-based_2024} was introduced to improve Swin Transformer V2's performance on CRC histology image classification, first, performing self-supervised pretraining and then fine-tuning with a layer-wise distillation technique on the NCT-CRC-HE-100K dataset. Furthermore, Guo et al. \cite{guo_predicting_2023} proposed a Swin Transformer workflow to identify CRC molecular biomarkers directly from WSIs. It was also used to predict microsatellite instability from a small dataset.

\subsection{Mamba for Cancer Classification}
Subsequently, SSM-based architectures \cite{gu2022efficientlymodelinglongsequences} like Vision Mamba \cite{zhu2024visionmambaefficientvisual} have been used to great effect in digital pathology tasks like WSI image classification \cite{zhang_2dmamba_2025}, due to their ability to capture global and local spatial context with far less computational power than transformers \cite{zheng_m3amba_2025,ding_combining_2025}. 2DMamba  \cite{zhang_2dmamba_2025} based on 2D selective SSM framework was introduced by integrating the 2D spatial structure of images into Mamba. Coupled with a hardware-aware optimized operator, it preserved spatial continuity and delivered strong computational efficiency. Furthermore, a memory-driven Mamba network, M3amba \cite{zheng_m3amba_2025}, was introduced to address vanilla Mamba's contextual forgetting when modeling long-range dependencies across thousands of instances. It consisted of a dynamic memory bank (DMB) that iteratively updated historical information, along with an intra-group bidirectional Mamba (BiMamba) block to improve feature representation and to fuse relevant historical information across groups, thereby facilitating richer inter-group connections. Another recent approach introduced a hybrid message-passing graph neural network (GNN) for local neighborhood interactions with Mamba to capture global tissue spatial relationships in WSIs \cite{ding_combining_2025}. It was validated for predicting progression-free survival in patients with early-stage lung adenocarcinoma (LUAD). Moreover, SlideMamba \cite{khan_slidemamba_2026}, an entropy-based adaptive fusion of GNNs and Mamba, was introduced to enhance representation learning in digital pathology. It provided a principled mechanism for combining complementary feature streams and improving multi-scale representation learning by emphasizing the branch with lower predictive entropy. Next, Vim4Path \cite{nasiri-sarvi_vim4path_2024}, a self-supervised vision mamba, was introduced for evaluation on the Camelyon16 dataset for both patch-level and slide-level classification. The model used a Vision Mamba architecture, inspired by state-space models in the DINO framework, for representation learning.

While this line of SSM-based research has shown promising results in histopathology and digital pathology, there are few investigations on neoplasias of the colon, and even fewer on the prophylaxis of colorectal carcinoma. Existing studies have largely focused on well-studied disease settings, such as lung and breast cancer. Mamba-based modeling for risk stratification of precancerous lesions (adenomatous polyps), such as TAs, remains an underexplored diagnostic need. To bridge this gap, this work investigates an ultra-lightweight mamba-based architecture for the classification of neoplastic tubular adenoma in CRC WSI data. It aims to leverage the efficiency of SSMs and their capacity for short- and long-range contextual modeling while concurrently capturing diagnostically relevant glandular and stromal patterns.



\subsection{Colorectal Cancer Histopathology Datasets}
High-quality datasets have played a central role in advancing automated CRC classification methods from histopathology images. Most datasets consist of hematoxylin and eosin (H\&E)-stained WSIs, considered the ``gold standard" for histopathology, that are divided into smaller patches and annotated by expert pathologists with tissue-type or prognostic labels at either the patch or slide level. While many studies use institution-specific, non-public datasets \cite{davri_deep_2022}, several publicly available datasets have been used for training and evaluation of CRC-related vision tasks. The NCT-CRC-HE-100K dataset comprises 100,000 labeled image patches from CRC tissue slides spanning nine colorectal tissue classes \cite{kather_100000_2018} and is widely used for supervised training. For tumor grading, Barbano et al. \cite{barbano_unitopatho_2021} introduced the UniToPatho dataset in 2021, consisting of nearly 10,000 image patches from six different classes corresponding to tissue type and dysplasia grade, including low- and high-grade dysplasia for TAs.

In contrast to these datasets, our Neoplastic Tubular Adenoma (NPTA) dataset \cite{ul-med}, \cite{rahu2026decodingfutureriskdeep} is designed as a retrospective cohort consisting solely of low-grade dysplastic TAs identified at screening colonoscopy. The dataset is released as an expertly annotated set of image patches derived from high-resolution WSIs of TA, stratified into two groups based on the outcomes of their follow-up screening(s): case and control. Case slides are derived from patients who subsequently developed CRC, and control slides are derived from patients who did not. This case–control design enables deep learning models such as XtraLight-MedMamba to learn subtle, early morphologic signals, perhaps associated with future CRC risk, from lesions that are not necessarily considered ``high-risk'' by current routine histologic guidelines.

\section{Methodology}
\begin{figure*}[h]
    \centerline{%
        \includegraphics[width=0.9\linewidth,height=7cm]{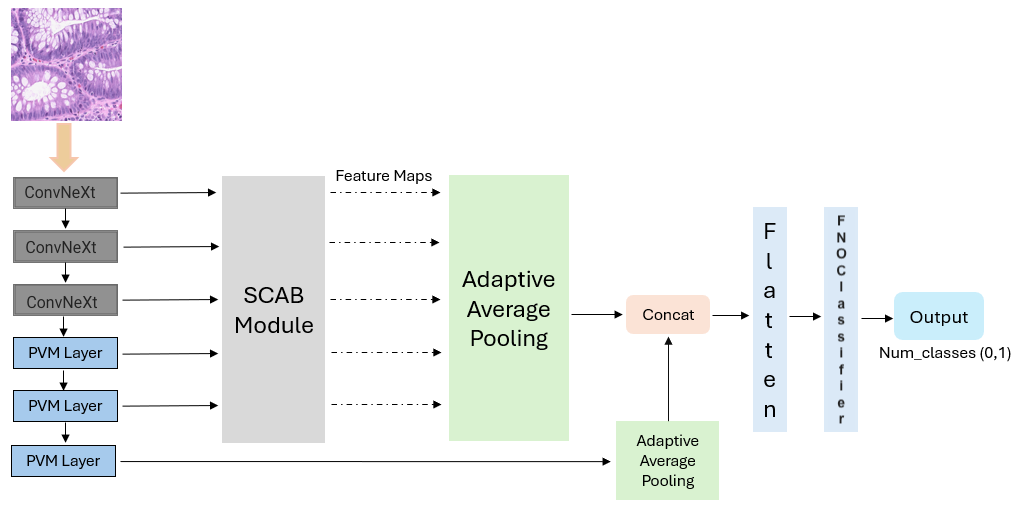}%
    }
    \caption{Architectural structure of XtraLight-MedMamba model for image classification task. The proposed architecture comprises ConvNeXt blocks for local morphological feature extraction, PVM layers for parallel state-space modeling, spatial and channel attention as SCAB modules, and an FNOClassifier to enforce fixed, nonnegative, orthogonal decision boundaries.}
    \label{fig2}
\end{figure*}

XtraLight-MedMamba adopts a convolutional neural network (CNN)-like backbone framework, replacing traditional convolution-only feature extractors with ConvNeXt blocks in the early stages and Parallel Vision Mamba (PVM) layers in the later stages as its primary feature extractors. The overall architecture is organized into six sequential stages, with the number of channels set as \([8,16,24,32,48,64]\) as shown in Fig. \ref{fig2} \cite{ul-med}. The first three stages employ ConvNeXt blocks \cite{woo2023convnextv2codesigningscaling} to generate shallow-to-mid level representations that capture semantically rich features from WSI tiles.
Each ConvNeXt block adopts a modernized convolutional design that replaces traditional residual bottlenecks with depthwise separable convolutions, Layer Normalization (LN), and a GELU-activation–based MLP-like expansion with a 4:1 channel ratio. 
These design choices, inspired by Transformer architectures, enhance feature expressivity and optimization stability while preserving the strong spatial inductive bias characteristic of convolutional networks.
The deeper layers, stages four to six, introduce PVM layers \cite{wu2024ultralightvmunetparallelvision}, \cite{ul-med} that act as state-space sequence mixers, enabling the modeling of long-range dependencies and information integration across spatial regions without the quadratic scaling of self-attention. This hierarchical combination enables the ConvNeXt front-end to capture localized morphological patterns, while the PVM back-end contextualizes them globally, bridging low-level texture information with high-level structural understanding, which is critical for differentiating neoplastic from non-neoplastic tissue patterns.

The multi-level features extracted from the three ConvNeXt blocks, together with the intermediate features from the PVM layers, are fed into a SCAB (spatial and channel attention bridge) module. To standardize the spatial dimensions for classification, feature maps from SCAB and the final PVM layer (stage 6) are passed through adaptive average pooling to achieve a common spatial resolution. The pooled feature maps are then concatenated to form a unified representation that retains relevant information.
The resulting high-dimensional feature representation is subsequently flattened into a one-dimensional vector and fed into the FNOClassifier, which produces logits that map the learned embeddings into the final output space. In this study, the classifier outputs a two-class probability distribution corresponding to control and case.

\subsection{Components of XtraLight-MedMamba Model}
\subsubsection{ConvNeXt--Shallow Feature Extractor for XtraLight-MedMamba Model}
\begin{figure}[h!]
    \centerline{%
        \includegraphics[width=0.7\linewidth,height=11cm]{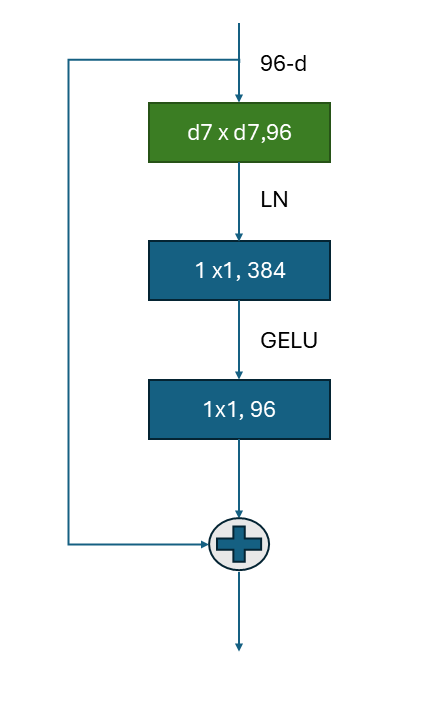}%
    }
    \caption{Unfolded ConvNeXt block for Shallow Feature Extraction in XtraLight-MedMamba Model.}
    \label{fig3}
\end{figure}

The ConvNeXt block \cite{article, Liu_2022_CVPR} builds upon the ResNet-style design by integrating modern architectural refinements inspired by Vision Transformer \cite{DBLP:journals/corr/abs-2010-11929}.
Each block begins with a \(7\times7\) depthwise convolution that captures the long-range spatial context within each channel while maintaining computational efficiency through channel grouping, as shown in Fig. \ref{fig3}.
The resulting feature map is then permuted from (B, C, H, W) to (B, H, W, C) to enable Layer Normalization and fully connected (linear) transformations that operate along the channel dimension, mimicking the feed-forward network structure of Transformers.
This “MLP” sublayer consists of two linear projections separated by a GELU activation, expanding the channel dimension by a factor of mlp\_ratio before projecting it back to the original dimensionality.
A learnable per-channel scaling parameter \(\gamma\) is applied to modulate the transformed output, followed by an inverse permutation that restores the tensor to (B, C, H, W).
Finally, a drop\_path is employed as a form of regularization before adding the residual connection, thereby preserving gradient flow and stabilizing training. This combination of depth-wise convolution, LayerNorm-based channel mixing, and residual learning enables ConvNeXt blocks to achieve high representational capacity and training stability with minimal computational overhead.

\subsubsection{Parallel Vision Mamba Module}
\begin{figure*}[h]
    \centerline{%
        \includegraphics[width=1.05\linewidth,height=7cm]{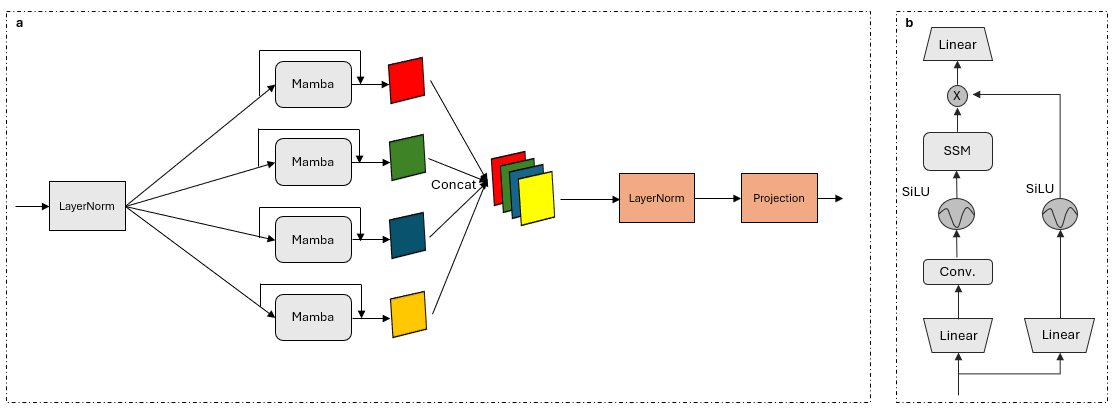}%
    }
    \caption{a) PVM layer in XtraLight-MedMamba for capturing of both short- and long-range spatial dependencies through parallel state space modeling b) Mamba module, a selective state space model for sequence modeling with linear computational complexity.}
    \label{fig:fig4}
\end{figure*}

The PVM module \cite{Sultana_Abouzahra_Asari_Aspiras_Liu_Sudakow_Cooper_2025, wu2024ultralightvmunetparallelvision}, as shown on the left side of Fig. \ref{fig:fig4} (part a) \cite{ul-med}, mainly consists of Mamba integrated with residual connections, which improve Mamba's ability to capture deep spatial relationships. Initially, the input \(Y_{in}^{C}\) with channel \(C\) goes through layer normalization. The features that are fed into Mamba are first branched out as \(X_1^{C/4}\), \(X_2^{C/4}\), \(X_3^{C/4}\), and \(X_4^{C/4}\), each with channel \(C/4\). The Mamba outputs, combined with the residual connection from the inputs and the optimization adjustment factor, are concatenated to obtain the four feature maps. The right side of Fig. \ref{fig:fig4} (part b) \cite{ul-med} shows the Mamba component used in the PVM layer. These features are concatenated to obtain \(Y_{out}\) with the number of channels as \(C\). The concatenated feature outputs are layer-normalized and then projected using a Projection operation. By splitting and processing deep features in parallel, the PVM module can capture multi-scale, intricate features at different granularities before fusion. This process significantly reduces the number of parameters by retaining the same receptive field, thereby avoiding increasing the channel dimension, which is a key consideration given Mamba's dependence on input dimensionality \cite{wu2024ultralightvmunetparallelvision}. The specific operations are expressed by the following equations:


\begin{align}
\label{eq:9}
X_{i}^{C/4} = S_{p}[\text{LN} (Y_{in}^{C})], \quad i = 1, 2, 3, 4
\end{align}
\begin{align}
\label{eq:10}
VM\_X_i^{C/4} &= Mamba(X_i^{C/4}) + \theta \cdot X_i^{C/4}, \quad i = 1, 2, 3, 4
\end{align}
\begin{align}
\label{eq:11}
Y_{\text{out}} = Cat\big(VM\_X_1^{C/4},\, VM\_X_2^{C/4},\,
VM\_X_3^{C/4},\, 
VM\_X_4^{C/4}\big)
\end{align}
\begin{align}
\label{eq:12}
\text{Out} = Proj\big[\text{LN}(Y_{\text{out}})\big]
\end{align}

Here, \(LN\) denotes the layer normalization, \(S_p\) is the split operation that splits the input into four separate branches, \(\theta\) is the residual scaling factor, \(cat\) denotes the concatenation operation where the outputs are concatenated and \(Proj\) is the projection operation where it projects concatenated representation back into the feature space.
 From Eq. \ref{eq:10}, we employed parallel Vision Mamba feature processing while keeping the total number of channels constant, thereby preserving high accuracy while achieving significant parameter reduction.
 
\subsubsection{SCAB Module}
Spatial and Channel Attention Bridge (SCAB) module is incorporated to refine intermediate representations and improve cross-scale feature integration and feature propagation \cite{ruan2022malunetmultiattentionlightweightunet, wu2024ultralightvmunetparallelvision}. SCAB module follows two paths: a spatial attention pathway that aggregates max-pooling, average pooling, concatenation operation, followed by extended convolution of shared weights, and a channel attention bridge pathway that includes global average pooling (GAP), concatenation operation, followed by fully connected layers (FCL), and sigmoid activation function. Together, this attention-based fusion improves the model's sensitivity, optimization stability, and multi-scale feature propagation \cite{wu2024ultralightvmunetparallelvision,Sultana_Abouzahra_Asari_Aspiras_Liu_Sudakow_Cooper_2025}.


\subsubsection{Fixed Non-Negative Orthogonal Classifier for Model Parameter Reduction}
The Fixed Non-Negative Orthogonal Classifier (FNOClassifier) \cite{kim2024fixed} module is a lightweight neural classifier that maps feature embeddings to class logits via structured, normalized projections. During initialization, the model partitions the input feature dimensions into approximately equal groups, each corresponding to an output class. A random permutation ensures that each class is assigned a unique subset of features, with any remaining dimensions randomly distributed among the classes to maintain balance. A binary base matrix is then constructed, where each row indicates the feature subset associated with a class, and is subsequently L2-normalized along the row dimension to enforce orthogonality and numerical stability. The model also includes an optional input feature normalization step, in which input vectors are scaled by their L2 norms, ensuring consistent magnitude across samples. In the forward pass, the input, which is either normalized or raw, is linearly projected using the scaled base matrix, modulated by a learnable scalar parameter \(W\) to produce class scores. This allows the FNOClassifier to serve as an efficient and interpretable projection-based classifier, emphasizing structured feature grouping and normalization for robust classification.

Fixed classifiers that incorporate orthogonality are recognized for their cost efficiency and, in some cases, their ability to outperform learnable classifiers on popular benchmarks. However, existing fixed orthogonal classifiers suffer from geometric limitations that fundamentally prevent them from invoking neural collapse \cite{kim2024fixed,XU2022251}.
\\

\noindent \textit{Inducing Neural Collapse through Fixed Non-Negative Orthogonal Classifier (FNOClassifier)}:

The FNOClassifier was proposed to resolve the issue that previous fixed orthogonal classifiers fail to invoke neural collapse (NC). NC is a critical phenomenon in which last-layer features converge to the simplex ETF (Equiangular Tight Frame) structure necessary to achieve global optimality in a layer-peeled model, due to their inherent geometric limitations \cite{kim2024fixed} \cite{XU2022251}. The development of the FNOClassifier relies on an analysis of zero-mean NC that explicitly accounts for orthogonality in non-negative Euclidean space. By satisfying the necessary properties of zero-mean NC, the FNOClassifier is designed to induce an optimal solution and maximize the margin of an orthogonal-layer peeled model. This classifier yields an inherent feature-dimension separation effect by mitigating feature interference in continual learning and addressing the limitations of mixup on the hypersphere in imbalanced learning, ultimately demonstrating significant performance improvements across various experiments.

\noindent \textbf{Fixed Non\mbox{-}Negative Orthogonal Classifier (FNOClassifier).}
Let $D$ and $F$ be the number of classes and feature dimension, respectively. A classifier with partial orthogonal weight matrix \(W\) is employed as 
$W \in \mathbb{R}^{F \times D}$, where the weights are not learned during training and require
\begin{align}
W^\top W &= I_D, \qquad W_{j,k} \ge 0 \ \ (\text{element\mbox{-}wise}), \label{eq:fnoc-constraints}
\end{align}
where \(I_D\) is identity matrix, \(W_{j,k}\) is the \((j,k)\)-th element of \(W\) and \(W^\top\) is the transpose of \(W\).
The orthogonality constraint ensures that each class has a distinct, independent direction in feature space, thereby maximizing inter-class separation and minimizing classifier-induced bias. The non-negativity constraint restricts the classifier to the positive orthant, promoting stable, consistent alignment between trained feature representations and their corresponding class vectors.

Based on an input feature representation, the logits produced by the classifier are defined by
\begin{align}
z(x) \;=\; \gamma\, W^\top x \quad (\gamma>0), \\
\qquad p(d \mid x)=\mathrm{softmax}_d\big(z(x)\big). \label{eq:fnoc-logits}
\end{align}
where \(z(x)\) is the logit vector, \(\gamma\) is a positive scaling parameter controlling the magnitude of the logits, and \(x\) is the input feature representation.
\(p(d|x)\) is the predicted probability of an input sample with feature representation where \(x\) belongs to class \(d\).
\\
\noindent \textbf{Neural collapse.} Neural collapse refers to a set of geometric properties that emerge toward the end phase of training neural networks using cross-entropy loss. First, intra-class collapse occurs. Under zero\mbox{-}mean neural collapse, features from the same class converge toward a shared class mean, where \(x_l^{(d)}\) is the feature embedding of the \(l\)-th sample belonging to the class \(d\), \(n_d\) and \(\mu_d\) are the number of samples and the class mean feature vector for class \(d\), respectively and defined as:
\begin{align}
\mu_d = \frac{1}{n_d} \sum_{l=1}^{n_d} x_l^{(d)}
\end{align}

\noindent The embeddings satisfy \eqref{intra} as the training progresses
\begin{align}
x_l^{(d)} &\xrightarrow[]{\text{training}} \mu_d 
\quad \text{(intra-class collapse)}
\label{intra}
\end{align}
 The convergence over the course of training is denoted by \(\xrightarrow[]{\text{training}}\).  The condition in \eqref{intra} indicates that within-class feature variance diminishes, and the same class feature embeddings concentrate around a single class.
 
 Second, class means centering occurs. Under zero-mean neural collapse, the weighted mean of the class becomes centered and converges to zero or the origin \cite{Papyan_2020}, \cite{jiang2023generalizedneuralcollapselarge}, \cite{chen2024neural}. Therefore, the features behave as:
\begin{align}
    \sum_{d=1}^D \psi_d \,\mu_d &= 0
\quad \text{(centered class means)}.
\label{eq:collapse-conds}
\end{align}
 In \eqref{eq:collapse-conds}, \(D\) is the total number of classes, the class probability is denoted by \(\psi_d\), where the global feature distribution is centered at the origin and the weighted average of all class means is zero, thus, ensuring a balanced and symmetric arrangement of class means in the feature space.

Finally, neural collapse enforces the equal\mbox{-}norm constraint on class means, which is given as:
\begin{align}
\|\mu_d\|=r, \qquad \forall_d,
\label{eq:equal-norm}
\end{align}
where all class means have the same Euclidean norm \(r\). This constraint ensures equal margins across classes in feature space by preventing dominance due to larger feature magnitudes. Integrated with intra-class collapse and centered class means, the resulting highly structured feature geometry concentrates samples from each class tightly around their respective means, where the class means are evenly distributed around the origin, and the angles between class directions are maximized. This yields compact, balanced, and maximally separated class representations consistent with Equiangular Tight Frame (ETF) geometry \cite{Papyan_2020}, \cite{zhu2021geometricanalysisneuralcollapse}.

\noindent \textbf{Class-Mean Optimization.} Under the NC theory, feature embeddings are represented at the level of class means rather than individual samples. Optimization of class mean representations $\mu_d$ becomes equivalent to optimization of individual feature vectors as intra-class variance diminishes. Consequently, training with a fixed classifier $W$
amounts to minimizing the cross\mbox{-}entropy over class means $\mu_d$:
\begin{align}
\min_{\{\mu_d\}}\left[\mathbb{E}_{d \sim \psi}\!\left[ -\log 
\frac{\exp\big(\gamma\, w_d^\top \mu_d\big)}{\sum_{m=1}^D \exp\big(\gamma\, w_m^\top \mu_d\big)} \right]\right]
\quad \text{s.t.} \quad 
\|\mu_d\|=r \quad  \label{eq:fnoc-obj} \ \
\end{align}
where \(d\) is the class index treated as a random variable obtained from the class prior \(\psi\) and $\mathbb{E}_{d \sim \psi}[\cdot] $ denotes the average over \(d\) under the class distribution \(\psi\). \(w_m\) is the weight vector of the competing class \(m\) and \(w_d\) is the $d$-th column of $W$ for the target class, where \(m \neq d\).
Under this formulation, optimization acts only on the feature extractor, aligning each class mean $\mu_d$ with its corresponding classifier while maintaining separation from the directions associated with other classes. When integrated with an equal-norm constraint in \eqref{eq:equal-norm}, it promotes a balanced and symmetric feature geometry consistent with the ETF structure of neural collapse \cite{Papyan_2020}.

\noindent \textbf{Margin Classification.} The objective in \ref{eq:fnoc-obj} is closely related to an implicit max-margin perspective.
In certain over-parameterized models, training with cross-entropy loss implicitly maximizes the classification margin between classes \cite{soudry2024implicitbiasgradientdescent}.
Therefore, the classification margin \cite{soudry2024implicitbiasgradientdescent} for class \(d\) versus class \(m\) is expressed in the following margin form:
\begin{align}
\max_{\{\mu_d\}}  \left[\min_{d \ne m}\big(w_d^\top \mu_d - w_m^\top \mu_d\big)\right]
\quad \text{s.t.}  \quad
\|\mu_d\|=r \quad \label{eq:collapse-conds1}
\end{align}
where the inner minimization ensures separation from the closest competing class. This drives each class means $\mu_d$ to align with its associated classifier direction $w_d$ while maintaining a distance from the other class directions. The equal-norm constraint ensures that min-margin and max-margin decision boundaries are consistent with NC.

\smallskip
\noindent\textbf{Relation to simplex\mbox{-}ETF.}
In standard learnable classifiers, NC theory predicts that classifier weights align with class means, i.e., $w_d \propto \mu_d$. The centered and normalized class means  form a simplex equiangular tight frame (ETF) $\langle \Tilde{\mu}_d,\Tilde{\mu}_{m}\rangle$ $\forall$  \(m, d \in \{1,..., D\}\) \cite{zhu2021geometricanalysisneuralcollapse}, \cite{Papyan_2020}:
\begin{align}
\left\langle \Tilde{\mu}_d, \Tilde{\mu}_{m} \right\rangle
=
\begin{cases}
1, & d = m, \\[4pt]
-\dfrac{1}{D-1}, & d \neq m.
\end{cases}
\end{align}

However, in FNOClassifier, $W$ is fixed with orthonormal, non\mbox{-}negative columns \cite{kim2024fixed}, satisfying~\eqref{eq:fnoc-constraints}. As a result, optimization of \eqref{eq:fnoc-obj} does not learn the classifier directions, but instead aligns the class means \(\mu_d\) with the fixed directions $w_d$ which are subject to zero-mean and equal-norm restrictions, producing a collapsed, balanced configuration without learning the partial orthogonal weight matrix $W$.

Although the non-negativity constraint precludes an exact simplex ETF inner-product structure, the resulting geometry still enforces
orthogonality of classifier directions, zero\mbox{-}mean centering of features, and balanced margins.

\section{Experimental Setup and Results}
\label{sec:exp}

\subsection{Dataset Curation}
\begin{figure}[h]
    \centering 

    \begin{subfigure}[b]{0.23\textwidth}
        \captionsetup{labelformat=empty}
        
        \includegraphics[width=\linewidth]{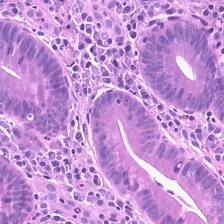}
         \\[0.8em]
         \includegraphics[width=\linewidth]{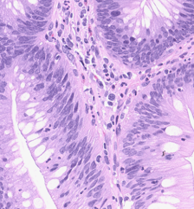}
        \caption{(a)}
    \end{subfigure}
    \hspace{0.004\textwidth} 
    \begin{subfigure}[b]{0.23\textwidth}
        \captionsetup{labelformat=empty}
    \includegraphics[width=\linewidth]{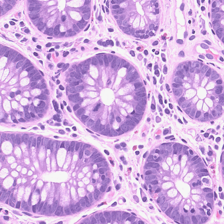}
     \\[0.8em]
        \includegraphics[width=\linewidth]{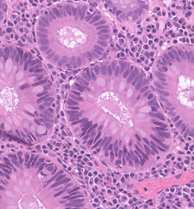}

    \caption{(b)}
    \end{subfigure}

    \caption{Sample images of the dataset used: (a) H\&E-stained WSI tiles from the case group consisting of tubular adenomas with low-grade dysplasia from patients who subsequently developed CRC (b)  H\&E-stained WSI tiles from the control group consisting of tubular adenomas with low-grade dysplasia from patients without subsequent development of CRC.}
    \label{fig:fig5}
\end{figure}

\begin{figure}[h]
    \centerline{%
        \includegraphics[width=0.9\linewidth,height=7cm]{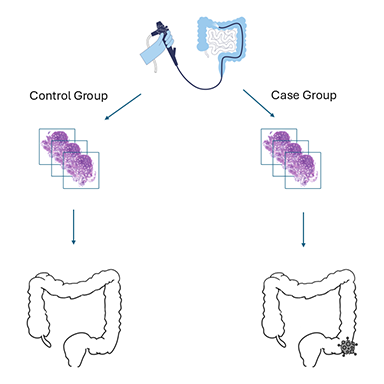}%
    }
    \caption{Illustration of post-colonoscopy outcomes.  Control Group: TA images from patients with no CRC development during follow-up. Case Group: TA images from patients who later developed CRC.}
    \label{fig6}
\end{figure}

A subset of WSIs was manually annotated by expert pathologists to mark Regions of Interest (ROIs) containing adenomatous epithelium. Three-color channel tiles of \(1024 \times 1024\) pixels containing tissue from the WSI were extracted. Each tile was labeled ROI-positive if it overlapped an annotated region, and ROI-negative otherwise. The tiles were resized to \(224 \times 224 \times 3\) to meet the input requirements of the EfficientNetV2S \cite{tan2021efficientnetv2smallermodelsfaster}. The model was trained on annotated WSI tiles to predict whether unseen tiles contained relevant tissue for this study.  

Following training, the model was applied to the full set of WSIs. These WSIs were tiled at \(1024 \times 1024 \times 3\) and then resized into smaller tiles of \(224 \times 224 \times 3\) pixels for model input as shown in Fig. \ref{fig:fig5}. During preprocessing, the trained EfficientNetV2S model evaluated each tile to determine whether it should be retained or discarded. The automatically generated ROIs were visually assessed for annotation accuracy. Tiles affected by quality issues, including tissue folding, edge artifacts, or suboptimal scan resolution, were excluded based on manual examination of the WSI patch location maps. Following curation, a total of 135,049 high-quality tiles were retained per class. The resulting dataset was subsequently split into training (70\%), validation (15\%), and testing (15\%) subsets \cite{rahu2026decodingfutureriskdeep}.


\subsubsection{Data Management Strategies}
Patients with low-grade TAs detected at screening colonoscopy were selected for the study after excluding individuals with known high-risk CRC predisposition or histologic features associated with advanced neoplasia. The cohort included 81 patients, comprising 41 males and 40 females, ranging in age from 54 to 95 years (average 70 years) whose biopsies showed TAs with low-grade dysplasia only and no histologic features suggestive of high-risk progression to CRC. The patients were grouped into case and control cohorts based on longitudinal outcomes. The case cohort included patients who subsequently developed CRC after screening colonoscopies in which low-grade TAs were identified, whereas the control cohort included patients without CRC development during follow-up despite detection of low-grade TAs on one or more examinations.
The control group had more biopsies and longer mean surveillance intervals, and patients from the case group were on average 6.86 years older than those in the control group.
Fig. \ref{fig6} shows the pictorial details of the two cohorts: the control and case groups \cite{rahu2026decodingfutureriskdeep}. For further details regarding the dataset, including patient inclusion/exclusion criteria and cohort definition, please refer to  \cite{rahu2026decodingfutureriskdeep} describing the study design.

\subsubsection{Data Acquisition}
Histological slides from both groups containing low-grade tubular adenomas were digitized using the same Leica Aperio AT2 \cite{AperioAT2} instrument, an automated whole-slide scanner capable of unattended scanning of up to 400 standard 75mm x 25mm glass slides through an integrated AutoLoader to generate image data for this study. The spatial resolution of the WSIs was 50,000 pixels per inch and  \(0.25\mu m/pixel\) at 40x. A Plan-Apochromat 20x/0.75 NA objective lens was used with an integrated 2x optical magnification changer, enabling 40x scanning with a single objective lens. Scanning was performed using a line-scan imaging method with fully automated tissue detection and automatic focus. The stored WSIs are 24-bit color images at gigapixel resolution, and the slides were generated in a contiguous pyramidal TIFF format, which was viewable through Aperio ImageScope software.

\subsection{Training Method}
All models, including transformer-based models, and mamba-based models with linear classifier and XtraLight mamba models were trained for 100 epochs under a common baseline configuration using Stochastic Gradient Descent (SGD) with a momentum of 0.99, learning rate of \(1\times10^{-5}\) and batch size of 256. The loss function was binary cross-entropy. Salt and pepper noise was randomly injected into the latent feature space during training rather than being directly applied to the input images to improve robustness and mitigate overfitting. The One-Cycle Learning Rate (OneCycleLR) scheduler, along with Stochastic Weight Averaging (SWA) to stabilize training, was incorporated to improve optimization stability.
All experiments were implemented in PyTorch and executed on a NVIDIA GeForce RTX Titan GPU. 

\subsection{Results}

\begin{table}[h!]
    \begin{center}
        \caption{Quantitative performances of Transformer-based models, mamba-based models with linear and fixed non-negative orthogonal classifiers.}
        \label{tab:tab2}
        \resizebox{\linewidth}{!}{ 
        \begin{tabular}{lccccc}
            \toprule
            \textbf{Models} & \textbf{Accuracy} & \textbf{F1} & \textbf{Precision} & \textbf{Recall}\\
            \midrule
            \textbf{Transformer-Based Models}\\
            \midrule
            Vision Transformer & 89.84\% & 0.8920 & 0.9519 & 0.8392 \\
            Swin Transformer & 89.52\% & 0.8878 & 0.9548 & 0.8296 \\
            \midrule
            \textbf{Mamba-Based Models with}\\
            \textbf{Linear Classifier}\\
            \midrule
            Conv. based & 94.24\% & 0.9431  & 0.9324  & 0.9540 \\
            MBConv. based & 92.44\% & 0.9221  & 0.9512  & 0.8948  \\
            Fused MBConv. based & 93.67\% & 0.9388  & 0.9076  & 0.9703  \\
            ConvNeXt based & 96.50\% & 0.9651 & 0.9606  &0.9697\\
            \midrule
             \textbf{XtraLight Mamba Models}\\
              \midrule
            Conv. based & 93.64\% & 0.9367  & 0.9321 & 0.9414  \\
            MBConv. based & 93.96\% & 0.9117  & 0.9029  & 0.8913\\
            Fused MBConv. based & 94.24\% & 0.9416  & 0.9547  & 0.9289 \\
            \textbf{ConvNeXt based } & \textbf{97.18\%} & \textbf{0.9767}  & \textbf{0.9666} & \textbf{0.9717} \\
            (ours, XtraLight-MedMamba)\\
            \bottomrule
        \end{tabular}
        }
    \end{center}
\end{table}

Table \ref{tab:tab2} summarizes the quantitative performance of Transformer-based architectures, Mamba-based variants, and XtraLight Mamba-based variants, employing both linear and fixed non-negative orthogonal classifiers. Among the Transformer models, which primarily utilize self-attention mechanisms to capture long-range dependencies across image patches, the Vision Transformer model achieved an overall accuracy of 89.84\% (F1 = 0.8920, precision = 0.9519, recall = 0.8392) with 7.39M parameters, while the Swin Transformer attained a comparable accuracy of 89.52\% (F1 = 0.8878, precision = 0.9548, recall = 0.8296) using a more compact 598K parameters.

In comparison, the Mamba-based models built upon State Space Models (SSMs) that process sequential representations bidirectionally, demonstrated consistently higher performance with substantially fewer parameters. Within the linear classifier group, the ConvNeXt-based Mamba variant achieved the highest accuracy of 96.50\% (F1 = 0.9651, precision = 0.9606, recall = 0.9697), outperforming the convolutional, MBConv, and Fused MBConv variants. The incorporation of the SCAB module further enhanced feature propagation, which proved beneficial for image classification. As the SSM effectively models hidden states over time, it excels at capturing both long- and short-range dependencies within spatial representations.

Similarly, the proposed XtraLight Mamba models maintained strong performance while significantly reducing model complexity. Our model, ConvNeXt-based XtraLight-MedMamba, achieved the best overall results, reaching an accuracy of 97.18\% (F1 = 0.9767, precision = 0.9666, recall = 0.9717) with only 32K parameters, surpassing all other tested configurations. These findings highlight that integrating ConvNeXt-style inverted bottlenecks into the Mamba framework strikes an optimal balance between efficiency and accuracy, outperforming both the Transformer and other Mamba variants.

Table \ref{tab:tab3} presents the normalized confusion matrix for the proposed XtraLight-MedMamba model. The classifier achieved exceptionally high discriminative performance, correctly identifying 97.7\% of positive samples (true positives) and 96.7\% of negative samples (true negatives). Only a small fraction of the cases were misclassified, with 2.3\% false negatives and 3.3\% false positives. These results demonstrate the model's strong sensitivity and specificity, confirming its robustness and reliability in distinguishing between positive and negative classes. Strong diagonal dominance confirms effective feature learning and generalization.


\begin{table}[h!]
\centering
\renewcommand{\arraystretch}{1.4} 

\caption{Confusion matrix for XtraLight-MedMamba}
\label{tab:tab3}

\begin{tabular}{
    c
    @{\hspace{10pt}}|@{\hspace{10pt}}
    c
    @{\hspace{12pt}}
    c
}
 & \textbf{Predicted Positive} & \textbf{Predicted Negative} \\ \hline
\textbf{Actual Positive}
 & \cellcolor{green!35!white}\hspace{4pt}0.9770 (TP)\hspace{6pt}
 & \cellcolor{red!15!white}\hspace{4pt}0.0230 (FN)\hspace{6pt} \\

\textbf{Actual Negative}
 & \cellcolor{red!20!white}\hspace{4pt}0.0333 (FP)\hspace{6pt}
 & \cellcolor{green!30!white}\hspace{4pt}0.9667 (TN)\hspace{6pt} \\
\end{tabular}
\end{table}
\subsection{Model Interpretability and Visual Explanation}
\begin{figure}[h]
    \centering 

    \begin{subfigure}[b]{0.23\textwidth}
        \captionsetup{labelformat=empty}    
        \includegraphics[width=\linewidth]{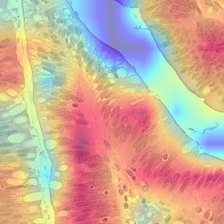}
         \\[0.8em]
        \includegraphics[width=\linewidth]{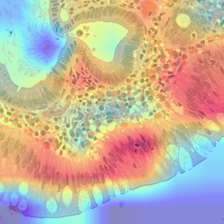}
            \\[0.8em]   
        \includegraphics[width=\linewidth]{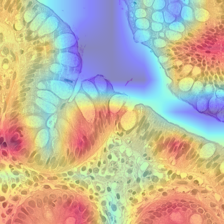}
        \caption{(a)}
    \end{subfigure}
    \hspace{0.005\textwidth} 
       \begin{subfigure}[b]{0.23\textwidth}
        \captionsetup{labelformat=empty}    
        \includegraphics[width=\linewidth]{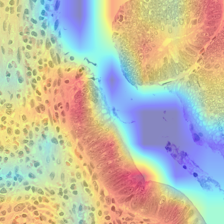}
         \\[0.8em]
        \includegraphics[width=\linewidth]{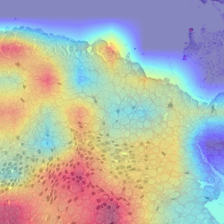}
              \\[0.8em] 
        \includegraphics[width=\linewidth]{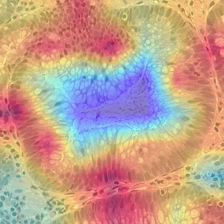}
        \caption{(b)}
    \end{subfigure}

    \caption{Grad-CAM based visualization of particular regions learned by the proposed XtraLight-MedMamba model. (a) Case samples highlight increased activation over the epithelium with subtle architectural and cytologic irregularities. (b) Control samples show diffuse attention across well-organized glandular structures, suggesting reliance on biologically relevant features. }
    
                            
    \label{fig:fig7}
\end{figure}
To aid interpretation of the decision-making behavior of the proposed XtraLight-MedMamba model, Gradient-weighted Class Activation Mapping (Grad-CAM) ~\cite{Selvaraju_2019} was used to visualize the spatial regions that contribute most to classification. As illustrated in Fig. \ref{fig:fig7}, the generated activation maps highlight the discriminative tissue areas that influenced the model’s predictions. Regions with warmer colors (red–yellow) correspond to areas of greater interest, whereas cooler colors (blue–purple) indicate lesser contribution. The Grad-CAM overlays demonstrate that the network predominantly focuses on histologically relevant structures—such as atypical epithelial glands, nuclear crowding, and stromal interfaces, rather than background artifacts. This alignment between activation patterns and pathologically relevant structures supports interpretability and reliability, confirming that XtraLight-MedMamba learns clinically meaningful representations rather than false correlations.

A closer look at the Grad-CAM visualizations in Fig. \ref{fig:fig7} shows that the model’s attention is drawn mainly to nuclear architecture. In both case and control images, the highlighted regions consistently align with areas rich in nuclear material, especially where the nuclei appear pseudostratified or show subtle irregularities in their basal orientation. In case samples, these activations often correspond to regions with more pronounced pseudostratification and occasional loss of polarity, whereas control samples display well-aligned nuclei with layering comparable to case nuclei. Together, these observations suggest that XtraLight-MedMamba differentiates tissues primarily by recognizing patterns of nuclear arrangement and polarity the same microscopic cues pathologists use to identify epithelial atypia and assess disease progression. 

\subsection{Model Parameters}
\begin{table}[h!]
    \begin{center}
        \caption{Model parameter comparison for Transformer-based models, mamba-based models with linear and fixed non-negative orthogonal classifiers.}
        \label{tab:tab1}
        \begin{tabular}{lc}
            \toprule
            \textbf{Transformer-Based Models} & \textbf{Model Parameters} \\
            \midrule
            ViT & 7,398,785 \\
            Swin Transformer & 598,099 \\
            \midrule
            \textbf{Mamba-Based Models with Linear Classifier} & \textbf{Model Parameters}\\
            \midrule
            Conv. Based & 49,641 \\
            MBConv. Based & 59,580  \\
            Fused MBConv. Based &  57,407  \\
            ConvNeXt Based & 53,385 \\
            \midrule
             \textbf{XtraLight Mamba Models} & \textbf{Model Parameters}\\
            \midrule
            Conv. Based & 28,329 \\
            MBConv. Based &  38,268 \\
            Fused MBConv. Based &  36,095  \\
            ConvNeXt Based (ours, XtraLight-MedMamba) & 32,073 \\
            \bottomrule
        \end{tabular}
    \end{center}
\end{table}

Table \ref{tab:tab1} summarizes the total number of trainable parameters across Transformer-based, Mamba-based, and XtraLight Mamba-based architectures. Among Transformer models, the Vision Transformer (ViT) had the highest parameter count, at approximately 7.4 million, followed by the Swin Transformer with 598 thousand parameters, owing to their hierarchical architectures. In contrast, Mamba-based architectures demonstrated substantially lower complexity, with parameter counts ranging from 49,000 to 59,000 for linear classifier variants. Further reductions were observed in the proposed XtraLight Mamba variants, with parameter counts reduced by nearly 40–50\% compared to the baseline Mamba models. The Conv.-based XtraLight Mamba model, with 28,329 parameters, had the fewest parameters among the models. Our ConvNeXt-based model, XtraLight-MedMamba, has the second-lowest count of the models studied, at only 32,073 parameters. This underscores the effectiveness of our proposed XtraLight design in minimizing computational cost while maintaining strong representational performance.

\subsection{Ablation Studies and Discussion}
F-1 Score and Recall were used for the ablation studies due to their robustness to class imbalance and clinical relevance. In this study, recall, also known as sensitivity, quantifies the model's ability to correctly identify high-risk adenomas, making false negatives particularly harmful. F1-score mitigates the potential class imbalance by providing a balanced assessment of precision and recall. Although accuracy provides completeness, F1-score and recall offer more clinically relevant evaluation and a more informative assessment of discriminative performance when comparing architectural and optimization variants.

\subsubsection{Effect of momentum on model performance}
\begin{figure}[h]
    \centerline{%
        \includegraphics[width=0.9\linewidth,height=7cm]{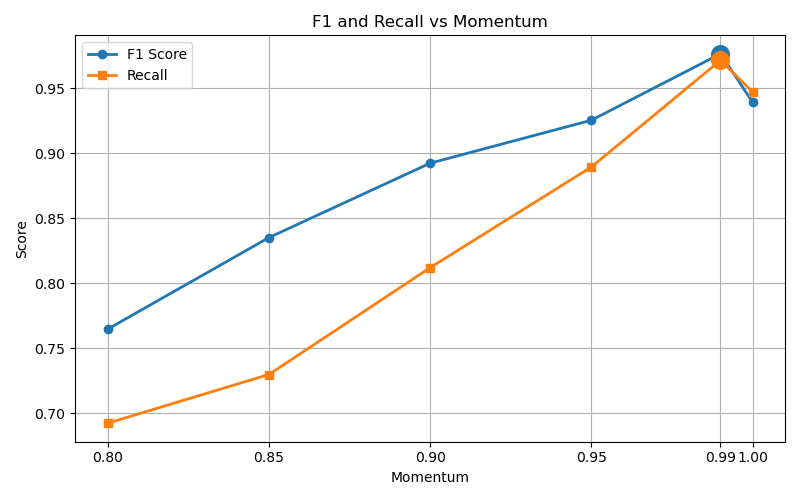}%
    }
    \caption{Analysis of momentum ranging from 0.80 to 1.0 versus F1-Score and Recall, where the momentum setting of 0.99 is highlighted, demonstrating the best overall performance among the evaluated configurations. }
    \label{fig9}
\end{figure}

An ablation study was conducted to assess the impact of momentum values 0.8, 0.85, 0.9, 0.95, 0.99, and 1.0 on model performance. As illustrated in Fig. \ref{fig9}, increasing momentum progressively improved the F1-Score and Recall, with optimal performance achieved at 0.99, followed by a slight dip at 1.0. By accumulating the exponential moving average (EMA) of past gradients, momentum smooths stochastic gradient updates. This enhances convergence along stable descent directions, reduces oscillations, and improves generalization ability of the model \cite{polyal}, \cite{Qian-Ning}, \cite{keskar}, \cite{pmlr-v28-sutskever13}. Although momentum at 1.0 shows descent performance, the slight dip compared to momentum at 0.99 suggests that reduced damping of gradient noise causes the optimizer to overshoot the optimal minimum at the extreme limit. Overall, high momentum effectively increases the step size in consistent directions, facilitating convergence toward flatter minima and improving generalization by escaping sharp local minima \cite{flat-minima}, \cite{keskar}.

\subsubsection{Effect of learning rate on model performance}
\begin{figure}[h]
    \centerline{%
        \includegraphics[width=0.9\linewidth,height=7cm]{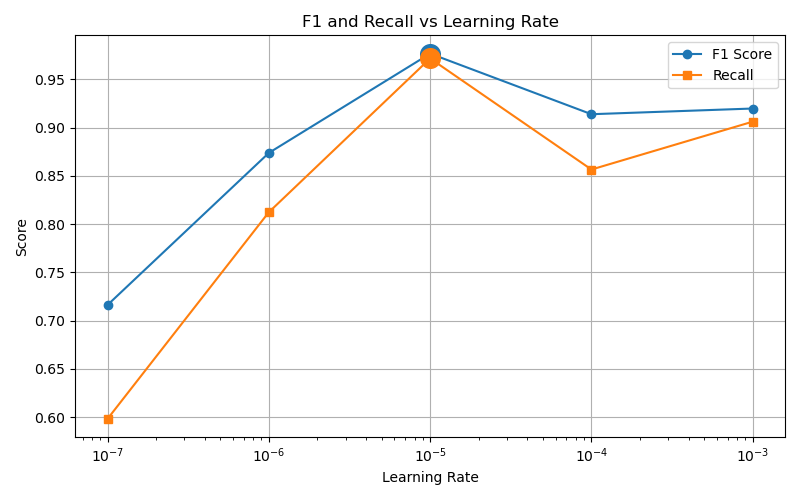}%
    }
    \caption{Analysis of learning rate ranging from \(1\times10^{-7}\) to \(1\times10^{-3}\) versus F1-Score and Recall, where learning rate setting of \(1\times10^{-5}\) is highlighted, demonstrating the best overall performance among the evaluated configurations.}
    \label{fig10}
\end{figure}

Impact of adjusting the learning rate from \(1\times10^{-7}\) to \(1\times10^{-3}\) on F1-Score and Recall is shown in Fig. \ref{fig10}.
Improvement in F1-Score and Recall was seen as the learning rate was increased from \(1\times10^{-7}\) to \(1\times10^{-5}\), with optimal performance achieved at \(1\times10^{-5}\). At lower learning rates, performance is severely constrained, leading to slower convergence and insufficient parameter updates caused by suboptimal optimization \cite{bengio2012practicalrecommendationsgradientbasedtraining}. Alternatively, increasing the learning rate beyond the optimal value can lead to overshooting of minima and unstable updates, degrading model generalization \cite{Goodfellow-et-al-2016}, \cite{keskar}. The performance dip observed at \(1\times10^{-4}\) and marginal recovery at \(1\times 10^{-3}\) suggest optimization instability at larger update steps, where they destabilize convergence and hinder optimization. The peak performance achieved at \(1\times10^{-5}\) learning rate represents the optimal balance between effective step size and convergence precision required for the generalization ability of the model.

\subsubsection{Effect of optimizers on model performance}
\begin{figure}[h!]
    \centerline{%
        \includegraphics[width=0.9\linewidth,height=7cm]{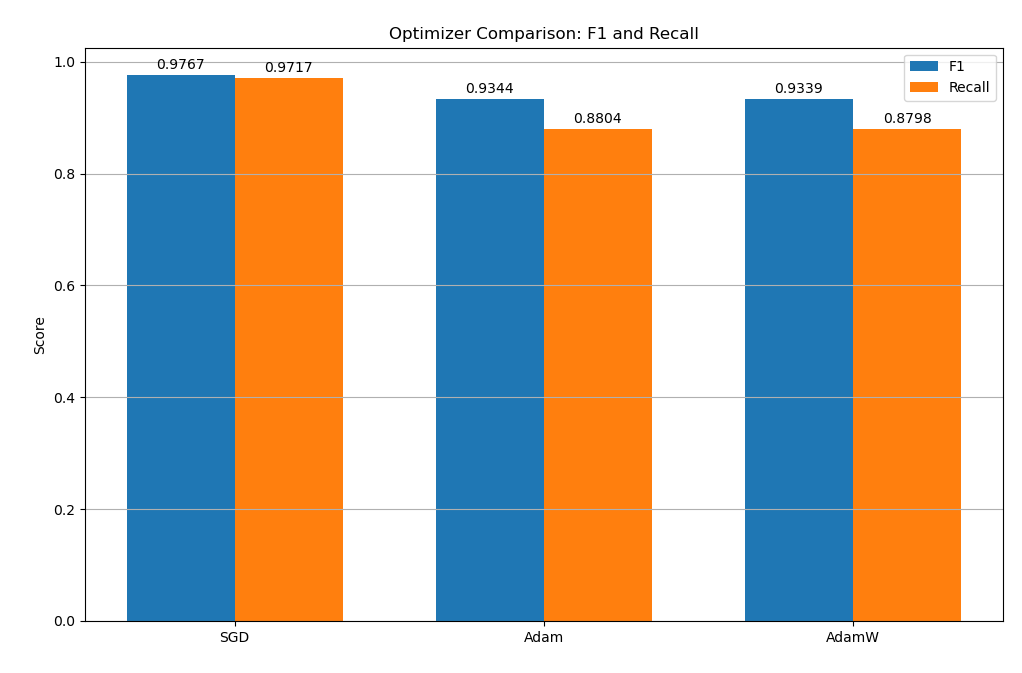}%
    }
    \caption{Performance comparison of SGD, Adam, and AdamW optimizers evaluated using F1-Score and Recall, where the SGD optimizer demonstrated improved classification performance.}
    \label{fig11}
\end{figure}

An ablation study was conducted to compare different optimizers, i.e., SGD, Adam, and AdamW, as shown in Fig. \ref{fig11}. Adaptive optimizers such as Adam accelerate convergence by adjusting per-parameter learning rates \cite{adam_iclr}, settling into sharper local minima, causing a generalization gap \cite{wilson2018marginalvalueadaptivegradient}, \cite{keskar}. AdamW, which decouples weight decay from gradient updates, improves regularization \cite{Loshchilov2017DecoupledWD} but did not outperform SGD based on our ablation experiments. SGD coupled with momentum often converges toward flatter minima, which are associated with improved generalization performance \cite{flat-minima}, thereby providing favorable optimization and generalization for histopathology classification.

\subsubsection{Effect of different classifier configurations on model performance}
An ablation study of model performance on different classifier configurations was conducted as shown in Table \ref{tab:tab4}. The configuration of the proposed model, XtraLight-MedMamba, with three linear and FNO layers, provides an optimal balance between structural regularization and flexibility. The all-FNO configuration imposes constrained or fixed orthonormal classifier directions, which may restrict the ability of the model to adapt heterogeneous patterns from feature representation, potentially limiting effective margin formation and sensitivity \cite{Papyan_2020}, \cite{zhu2021geometricanalysisneuralcollapse}. Alternatively, the all-hadamard \cite{9533441}, \cite{blank2020quantumclassifiertailoredquantum} configuration promotes the decorrelation of the feature and inter-class separability but lacks the structured geometric alignment introduced by FNO layers \cite{Saxe2013ExactST}.
The proposed hybrid design, consisting of three linear and two FNO layers, leverages the combination of adaptive linear layers to capture specific feature geometry under the orthogonality constraints imposed by the FNO layers, yielding improved feature representation, wider decision margins, and superior generalization performance.

\begin{table}[h]
    \begin{center}
        \caption{Performance comparison of different classifier configurations within the proposed framework.}
        \label{tab:tab4}
        \resizebox{\linewidth}{!}{ 
        \begin{tabular}{lcccccc}
            \toprule
            \textbf{Classifiers} & \textbf{Accuracy} & \textbf{F1} & \textbf{Precision} & \textbf{Recall}& \textbf{\# params}\\
            \midrule 
            All FNO & 80.95\% & 0.8392 & 0.7262 & 0.7938 & \textbf{25,884}\\
            3 Linear 2 Hadamard  & 87.77\% & 0.8884 & 0.8173 & 0.8730 & 32,401\\
            All Hadamard  & 92.44\% & 0.9252 & 0.9163 & 0.9342 & 26,260\\
            \textbf{XtraLight-MedMamba} & \textbf{97.18\%} & \textbf{0.9767}  & \textbf{0.9666} & \textbf{0.9717} & 32,073\\
            (ours, 3 Linear, 2 FNO)\\ 
            \bottomrule
        \end{tabular}
        }
    \end{center}
\end{table}

\subsubsection{Model parameters}
The number of trainable parameters for each classifier configuration is shown in Table \ref{tab:tab4}. Due to structural differences, the parameter sizes of the learnable linear classifier, FNOClassifier, and the Hadamard classifier differ. Configurations such as all-linear layers require more parameters due to fully trainable matrices, thereby expanding the model's representational capacity. FNO layers exhibit minimal additional trainable parameters due to their fixed orthonormal classifier directions. This imposes structured geometric constraints and reduces model complexity. Hadamard layers rely on structured transformations that produce comparatively fewer learnable parameters while enhancing feature decorrelation. The findings demonstrate that higher parameter complexity alone is insufficient to explain performance improvements. Although the proposed model, XtraLight-MedMamba (3-linear and 2-FNOClassifier) contains the second highest number of parameters at 32,073, it achieves the superior performance compared to other classifier configurations.

\subsection{Mismatched predicted outputs of XtraLight-MedMamba}
\begin{figure}[h!]
    \centering 

    \begin{subfigure}[b]{0.23\textwidth}
        \captionsetup{labelformat=empty}
        \includegraphics[width=\linewidth]{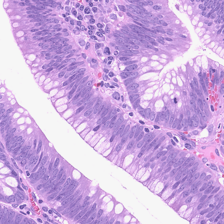}
         \\[0.8em]
        \includegraphics[width=\linewidth]{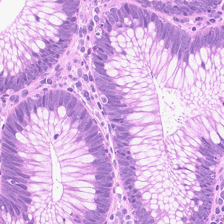}
         \\[0.8em]
        \includegraphics[width=\linewidth]{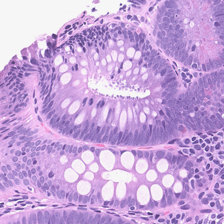}
        \caption{(a)}
    \end{subfigure}
    \hspace{0.005\textwidth} 
    \begin{subfigure}[b]{0.23\textwidth}
        \captionsetup{labelformat=empty}
        \includegraphics[width=\linewidth]{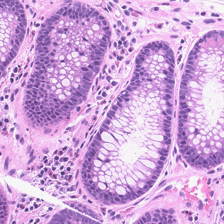}
         \\[0.8em]
        \includegraphics[width=\linewidth]{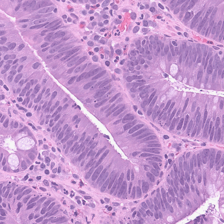}
         \\[0.8em]
        \includegraphics[width=\linewidth]{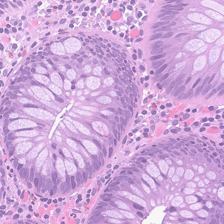}
        \caption{(b) }
        
    \end{subfigure}

    \caption{Representative mismatched predictions made by XtraLight-MedMamba. (a) Target: Case, Predicted: Control. The glands in these sections retain their architecture, with evenly spaced crypts and relatively consistent nuclear alignment. Cytologic atypia in these sections is subtle, with mild nuclear enlargement and crowding, morphological features that closely resemble those of low-risk tubular adenomas. The lack of overt glandular complexity or pronounced dysplasia likely contributed to the model misclassifying the sample as Control. (b) Target: Control, Predicted: Case. In contrast, these Control samples exhibit focal areas of gland crowding, mild loss of polarity, and mild pseudostratification of the nuclei with darker nuclei. These morphological patterns are more commonly associated with higher-risk lesions, which may have led the model to treat them as case-associated features.
    }
    
    \label{fig:fig8}
\end{figure}

In Fig. \ref{fig:fig8} (a), a misclassified case is illustrated, showing tiles that clearly demonstrate low-grade adenomatous features, including nuclear crowding with elongation and hyperchromasia, nuclear pseudostratification, and goblet cell depletion. This misclassification likely reflects the patchy nature of dysplasia and its presence within mixed histology, where focal dysplasia and adjacent benign crypts may obscure subtle neoplastic changes and dilute the morphological features on which the model relies. This finding highlights the need for finer-grained annotations, including tighter ROIs and more examples of early dysplasia during training.

In Fig. \ref{fig:fig8} (b), the tiles belong to the control group; however, the model misclassified them as the case group that progressed to CRC despite the largely preserved architecture. Histologically, the glands are orderly and well-spaced, crypts are well-formed and well-spaced, goblet cells are plentiful, and nuclei are basally aligned without convincing pseudostratification or high-grade cytologic atypia. There are no obvious morphologic features that would suggest progression toward CRC. One possibility is that the model is being influenced by non-biological cues, such as epithelial crowding near the tissue edge or darker nuclear staining, rather than true progression-related morphology.

\section{Conclusion}
In this work, we introduce XtraLight-MedMamba, an ultra-lightweight state-space model for classifying Neoplastic Tubular Adenomas (NPTA). The proposed architecture combines ConvNeXt blocks for shallow feature extraction with PVM layers to efficiently capture both short- and long-range dependencies while maintaining a minimal parameter footprint. The performance was further enhanced by integrating the SCAB module to improve cross-scale feature extraction and feature propagation. Subsequently, the FNOClassifier exhibited minimal additional trainable parameters due to their fixed orthonormal classifier directions, which imposed structured geometric constraints and reduced model complexity.
The proposed approach achieved superior classification performance on a curated dataset of low-grade tubular adenomas, outperforming both transformer-based and conventional Mamba architectures while using substantially fewer parameters. Grad-CAM analysis demonstrated that the network captured
subtle histological patterns, including nuclear architecture and  epithelial atypia, supporting both interpretability and clinical relevance, by highlighting discriminative features
that conventional methods may overlook.

\section*{Acknowledgment}

The authors acknowledge the South Bend Medical Foundation (SBMF) team for providing access to the Neoplastic Tubular Adenomas dataset.

\section*{References}
\vspace{-10pt}
\bibliography{lit, related_works}

\newpage
\section{Biography Section}
 \vspace{-40pt}
\begin{IEEEbiography}[{\includegraphics[width=1in,height=1.25in,clip,keepaspectratio]{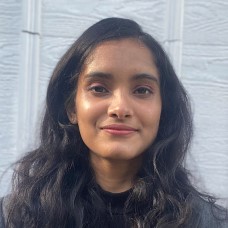}}]{Aqsa Sultana}
 (Student Member, IEEE) received the M.S. degree in Computer Engineering from the University of Dayton, Dayton, OH, USA, where she is currently pursuing the Ph.D. degree in Electrical Engineering. Her research interests include remote sensing application, neuromorphic computing and spiking neural networks, automated feature extraction, medical image analysis, and pattern recognition for oncology applications. She received the 2025 IEEE Dayton Section Krishna M. Pasala Memorial Scholarship in recognition of her academic excellence.
\end{IEEEbiography}
\vspace{-40pt}

\begin{IEEEbiography}
[{\includegraphics[width=1in,height=1.25in,clip,keepaspectratio]{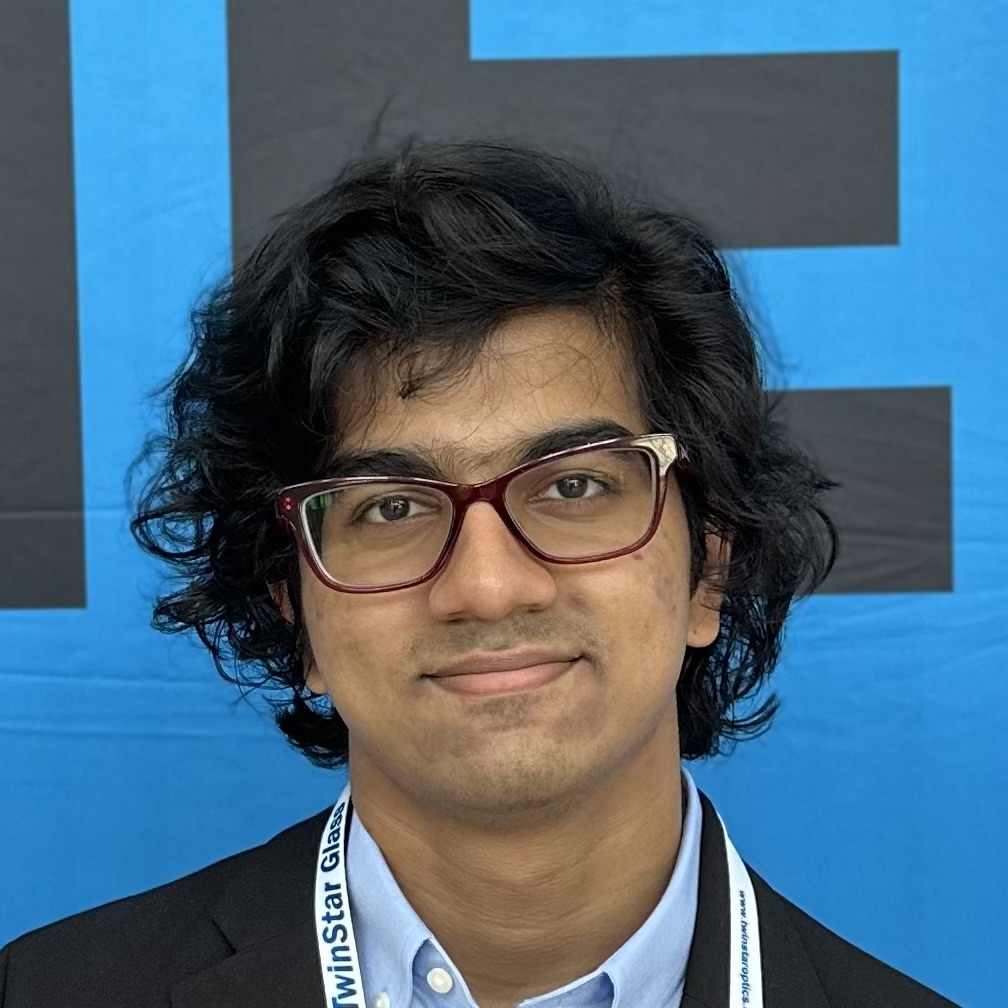}}]{Rayan Afsar}
 (Student Member, IEEE) is currently pursuing his Bachelor's degree in Computer Science at the University of Georgia. His research interests focus on integrating computer vision with remote sensing to support human-based decision-making and to validate environmental models for understanding Earth systems.
\end{IEEEbiography}
\vspace{-40pt}

\begin{IEEEbiography}
[{\includegraphics[width=1in,height=1.25in,clip,keepaspectratio]{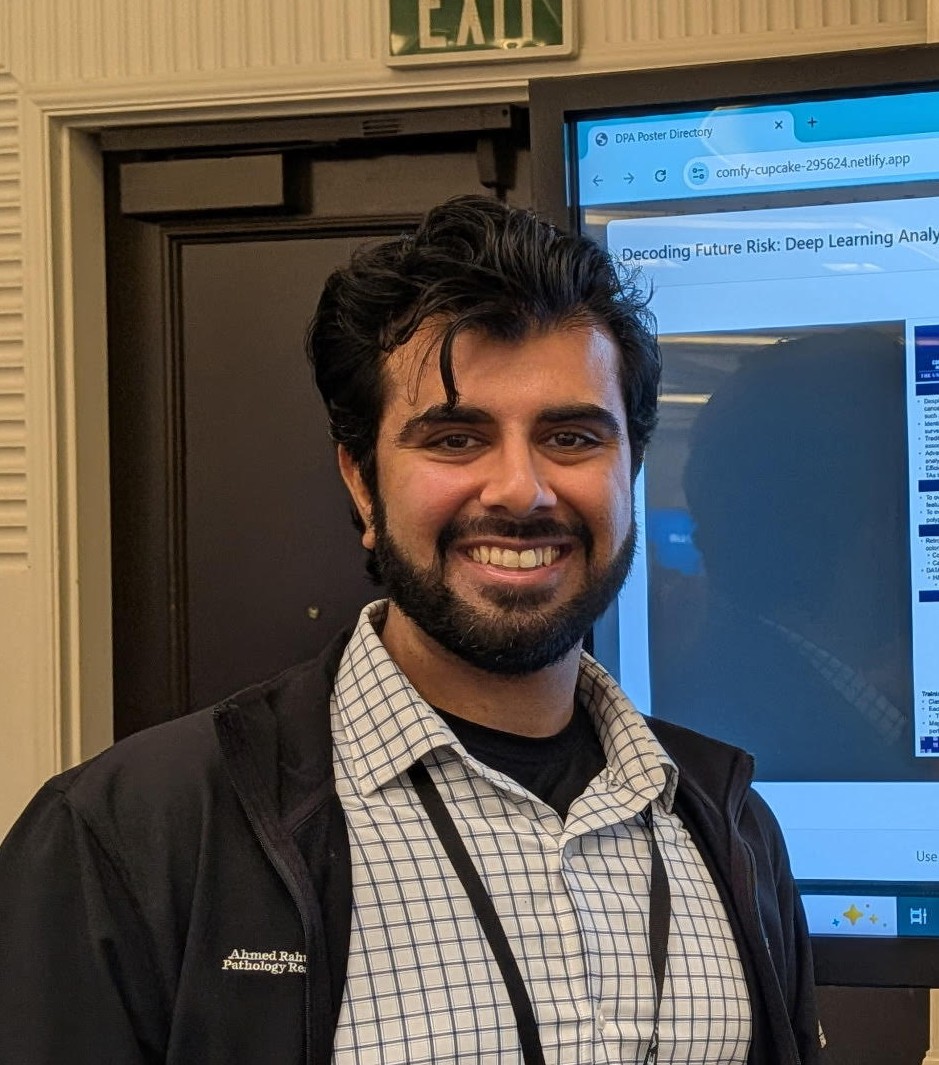}}]{Ahmed Rahu, \textit{MD}} received his B.S. in Biomedical Engineering from George Mason University, Fairfax, VA, and earned his M.D. from Ross University School of
Medicine. He is currently a PGY-2 Pathology resident at the University of Toledo Medical Center. His academic and clinical interests center on bridging
technology and medicine to advance disease prevention, diagnostic precision, and therapeutic innovation. His research focuses on digital pathology,
computational pathology, and machine-learning–based image analysis, with particular interest in risk stratification of pre-malignant lesions, predictive modeling in
oncologic pathology, and the development of lightweight, interpretable deep learning architectures for clinical deployment.
\end{IEEEbiography}
\vspace{-40pt}

\begin{IEEEbiography}
[{\includegraphics[width=1in,height=1.25in,clip,keepaspectratio]{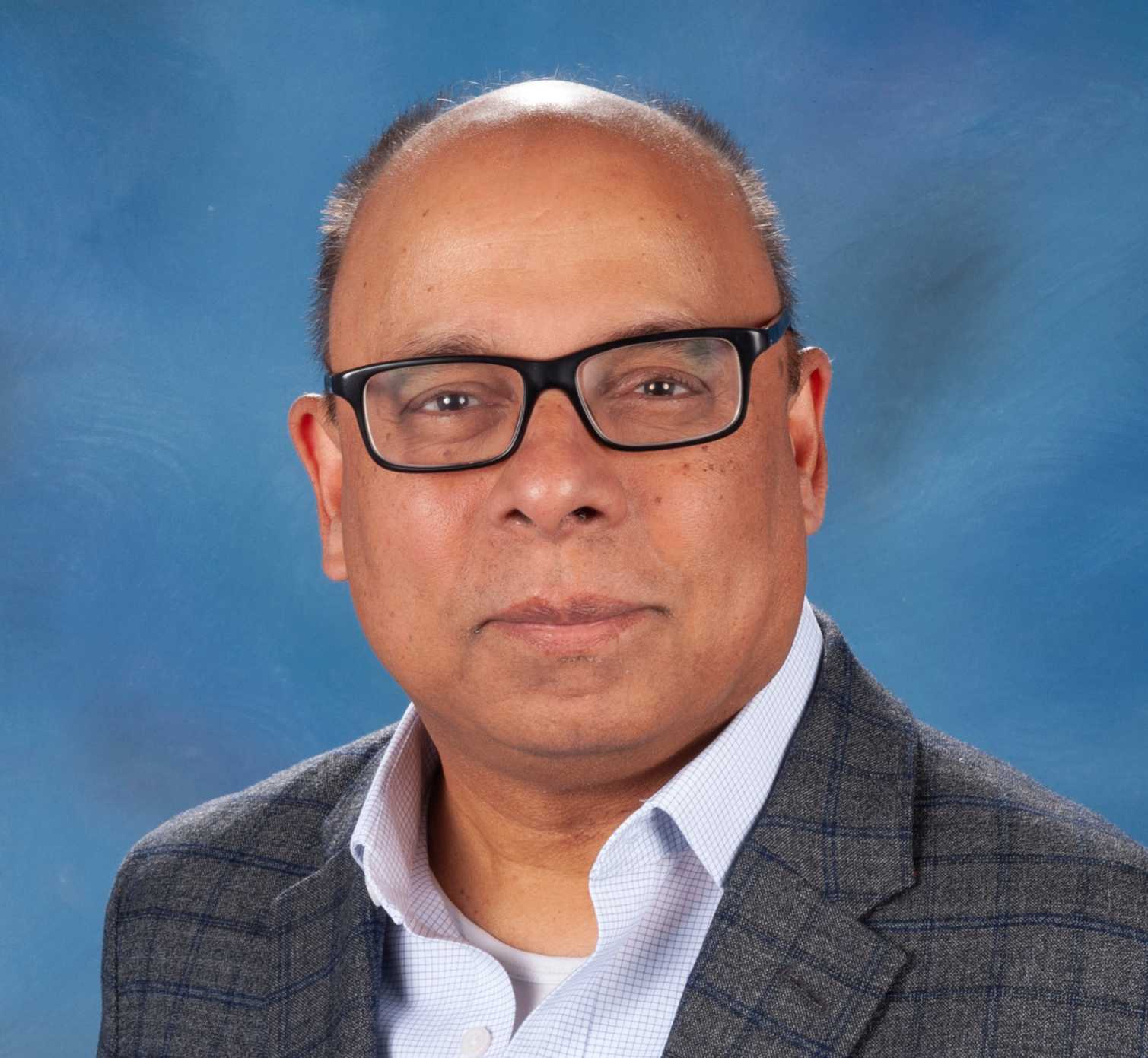}}]{Surendra P. Singh, \textit{MD}} joined the Consultants in Laboratory Medicine in 2000 and had various roles
including Medical Director/Vice Chairman in Anatomic Pathology, Microbiology, Medical Education, and
ProMedica BioRepository. He is board-certified in anatomic and clinical pathology. He completed a
Gastrointestinal pathology fellowship at Harvard Medical School /Beth Israel Deaconess Medical Center,
Boston. He completed his Anatomic and Clinical Pathology residency and a Surgical Pathology fellowship
at the Ohio State University. He also completed an American Cancer Society Fellowship in Clinical
Oncology at The Ohio State University. He received his graduate M.B; B.S degree and postgraduate M.D
degrees from the M.S University and Medical College of Baroda, India.
Dr. Singh maintains a faculty appointment at the College of Medicine and Life Sciences, University of
Toledo as a Clinical Associate Professor.
He also chairs the Aurora GI and Liver Council. He has authored several articles in peer-reviewed journals
and has received numerous awards, including the Stowell Orbison Award.
His primary interests include general surgical, gastrointestinal, liver, and oncologic pathology,
immunohistochemistry, microbiology, infectious pathology, and medical education.
\end{IEEEbiography}
\vspace{-40pt}

\begin{IEEEbiography}
[{\includegraphics[width=1in,height=1.25in,clip,keepaspectratio]{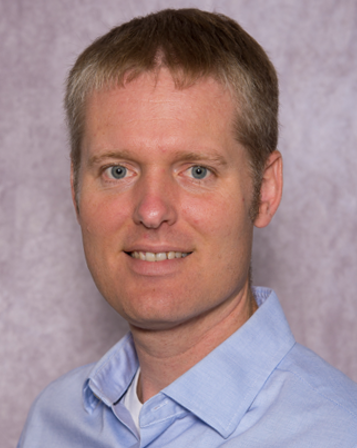}}]{Brian Shula}  is a Lead Mechanical Design Engineer for Aircraft Wheels and Brakes at Honeywell, with over 20 years of structural numerical simulation experience, predominantly in the aerospace industry.  Brian has applied machine learning tools in structural simulation settings by developing finite element surrogate models to facilitate design space exploration.  Mr. Shula earned his BSME and MSME from the University of Notre Dame and holds a Professional Engineer license in Ohio.
\end{IEEEbiography}
\vspace{-40pt}

\begin{IEEEbiography}
[{\includegraphics[width=1in,height=1.25in,clip,keepaspectratio]{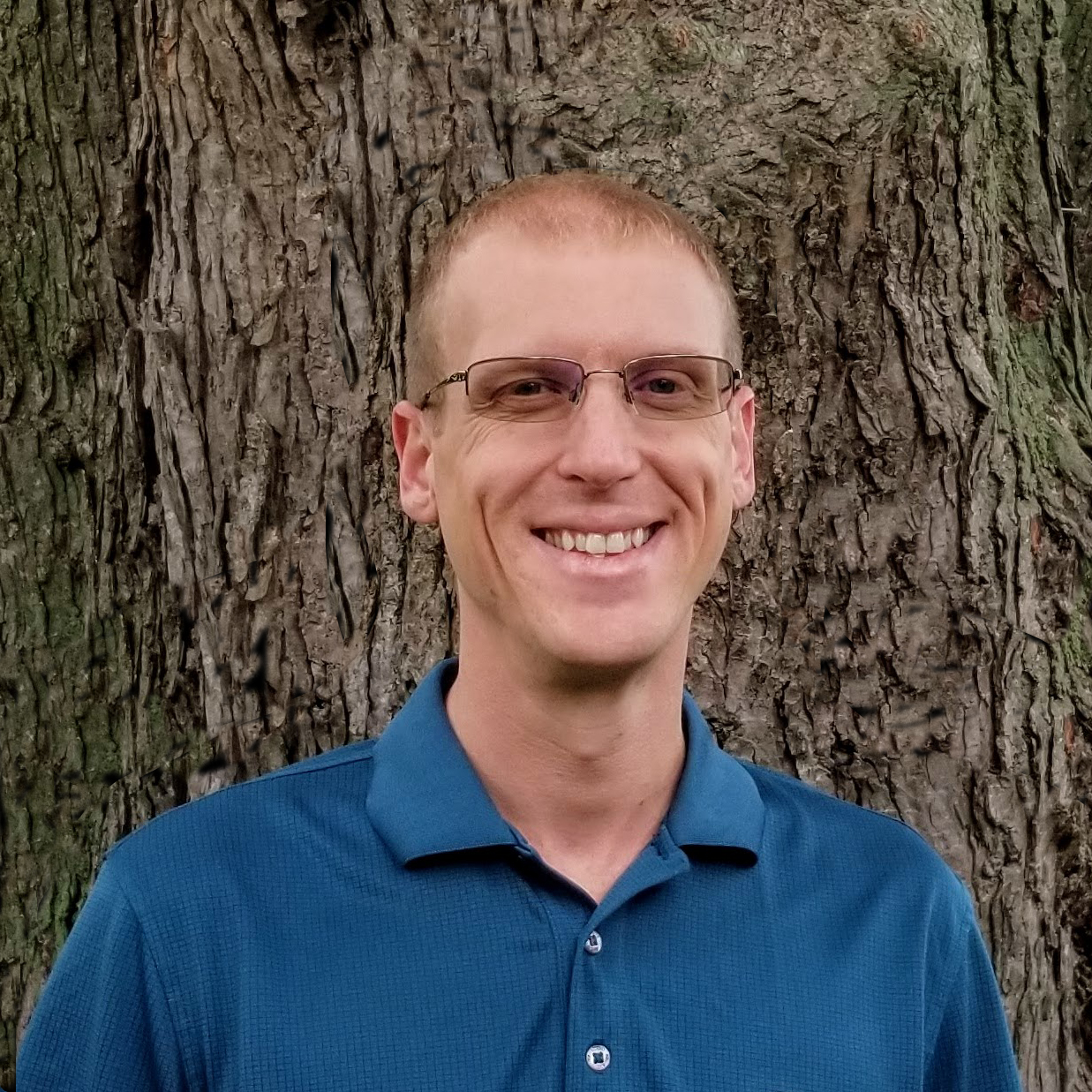}}]{Brandon Combs} is a Cisco-certified Technical Solutions Architect at the South Bend Medical Foundation, where he has supported clinical laboratory operations and advanced digital pathology systems for over a decade, beginning as an independent contractor. He received a Bachelor of Science degree in Mathematics and Physics from Indiana University, followed by a Master of Science degree in Applied Mathematics and Computer Science.

His technical and research interests include computational pathology, automated image-quality assessment, machine-learning–based quality control, clinical workflow optimization, and scalable cloud-based healthcare systems. His professional experience spans software engineering, system architecture, and enterprise network administration for distributed clinical environments.
\end{IEEEbiography}
\vspace{-90pt}

\begin{IEEEbiography}
[{\includegraphics[width=1in,height=1.25in,clip,keepaspectratio]{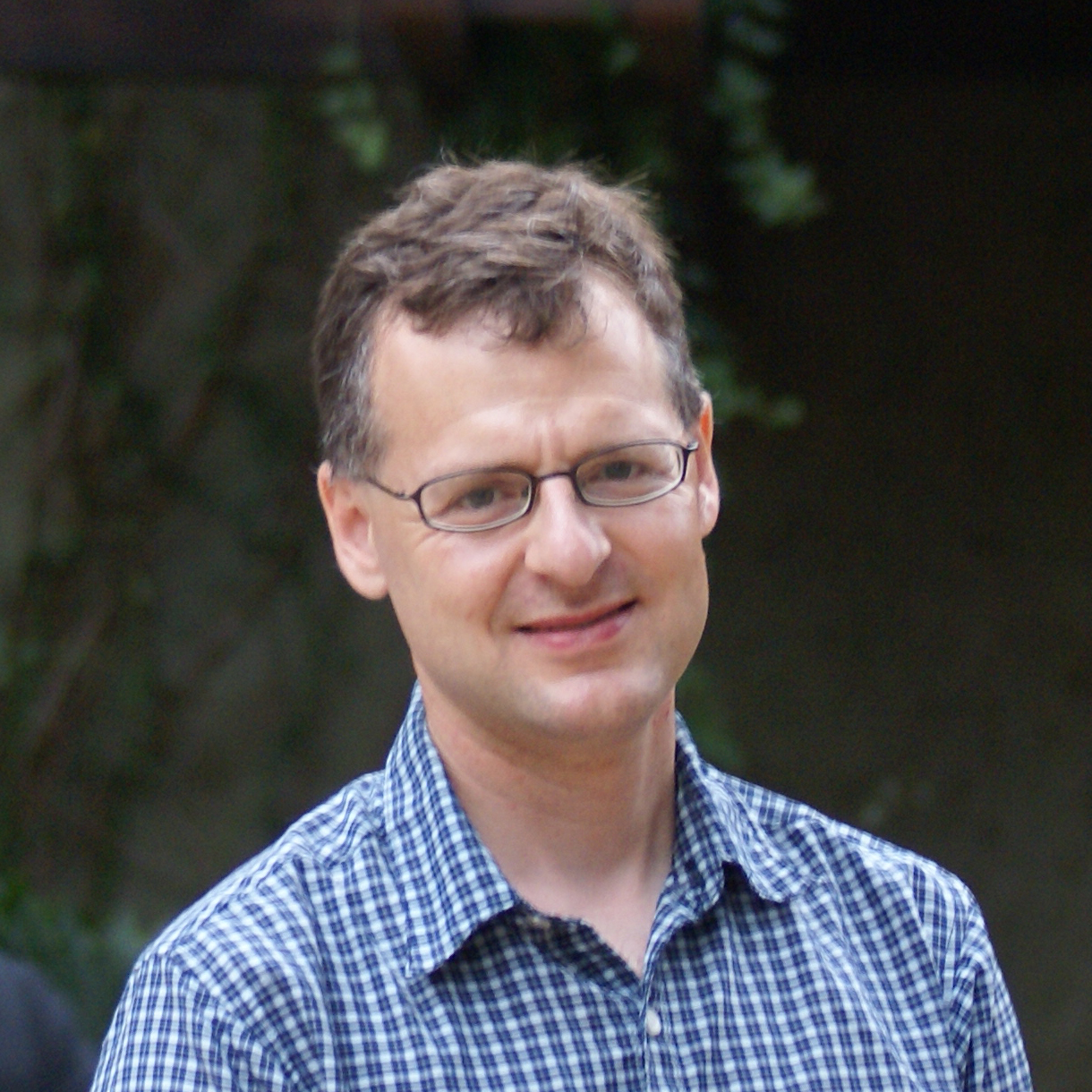}}]{Derrick Forchetti, \textit{MD}} received a B.A. degree (cum laude) in chemistry and German from Wabash College in 1993, an M.D. degree from Indiana University School of Medicine in 1997, and an M.S. degree in data science from the University of Wisconsin Extended Campus in 2021. He completed a five-year residency training program in anatomic and clinical pathology at Ball Memorial Hospital in 2002.

He is a board-certified anatomic and clinical pathologist with nearly 25 years of practice experience, with additional board certification in clinical informatics. He currently serves as a pathologist at South Bend Medical Foundation and as a volunteer Assistant Professor at the University of Toledo College of Medicine and Life Sciences. He is a member of the College of American Pathologists Digital and Computational Pathology Committee. His research interests include automated quality control for whole slide imaging, digital pathology workflows, and the application of artificial intelligence in diagnostic laboratories.
\end{IEEEbiography}
\vspace{-90pt}

\begin{IEEEbiography}
[{\includegraphics[width=1in,height=1.25in,clip,keepaspectratio]{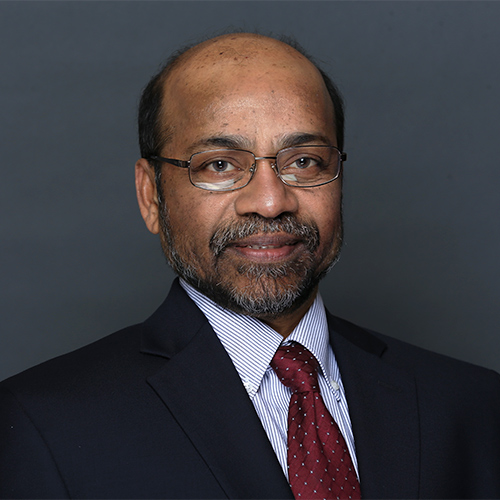}}]{Vijayan K. Asari, \textit{PhD}}
(Senior Member, IEEE) Received
the Ph.D. degree in electrical engineering from the
Indian Institute of Technology Madras, Chennai, India, in 1994. He is currently a Professor of Electrical and Computer Engineering and the Ohio Research Scholars
Endowed Chair in wide area surveillance with the University of Dayton, Dayton, OH, USA, where he is also the Director of the Center of Excellence for Computational Intelligence and Machine Vision (Vision Lab). He holds five U.S. patents and has published more than 800 research articles, including 147 peer-reviewed journal papers and 7 edited books co-authored with his students, colleagues, and collaborators in the areas of image processing, pattern recognition, machine learning, deep learning, and artificial
neural networks. Dr. Asari is a recipient of several teaching, research, advising, and technical leadership awards, including the Outstanding Engineers and Scientists Award for Technical Leadership from The Affiliate Societies Council of Dayton in April 2015, the Sigma Xi George B. Noland Award for Outstanding Research in April 2016, and the University of Dayton School of Engineering Vision Award for Excellence in August 2017. Dr. Asari was selected as a Fulbright Specialist by the US Department of State’s Bureau of Educational and Cultural Affairs (ECA) and World Learning in 2017.  He was also selected as the European Union's Erasmus+ Faculty Fellow in 2018. Professor Asari is a Senior Member of IEEE and an elected Fellow of SPIE, and a Co-Organizer of several IEEE and SPIE conferences and workshops.  
\end{IEEEbiography}

\end{document}